%% file: main.tex
\documentclass[10pt,twocolumn,letterpaper]{article}
\usepackage[pagenumbers]{iccv}
\input{preamble}
\definecolor{iccvblue}{rgb}{0.21,0.49,0.74}
\usepackage[pagebackref,breaklinks,colorlinks,allcolors=iccvblue]{hyperref}

\usepackage{algorithm}
\usepackage{listings}
\usepackage{etoolbox}
\makeatletter
\AfterEndEnvironment{algorithm}{\let\@algcomment\relax}
\AtEndEnvironment{algorithm}{\kern2pt\hrule\relax\vskip3pt\@algcomment}
\let\@algcomment\relax
\newcommand\algcomment[1]{\def\@algcomment{\footnotesize#1}}
\renewcommand\fs@ruled{\def\@fs@cfont{\bfseries}\let\@fs@capt\floatc@ruled
  \def\@fs@pre{\hrule height.8pt depth0pt \kern2pt}%
  \def\@fs@post{}%
  \def\@fs@mid{\kern2pt\hrule\kern2pt}%
  \let\@fs@iftopcapt\iftrue}
\makeatother

\title{Scalable Image Tokenization with Index Backpropagation Quantization}
\author{
\hspace{-3mm}
\noindent
\textbf{Fengyuan Shi}$^{1,3*}$\quad
\textbf{Zhuoyan Luo}$^{2,3 *}$\quad
\textbf{Yixiao Ge}$^{3\dagger\text{\Letter}}$\quad
\textbf{Yujiu Yang}$^{2}$\quad
\textbf{Ying Shan}$^{3}$\quad
\textbf{Limin Wang}$^{1\text{\Letter}}$\quad\\
\textbf{\small $^*$equal contribution}\quad \textbf{\small $^\dagger$ project lead}\quad \textbf{\small $^\text{\Letter}$ corresponding author}\\~\\
$^1$Nanjing University\quad
$^2$Tsinghua University\quad
$^3$ARC Lab, Tencent PCG\quad
}

\begin{document}

\twocolumn[{
\maketitle
\centering
\vspace{-10mm}
\includegraphics[width=0.99\linewidth]{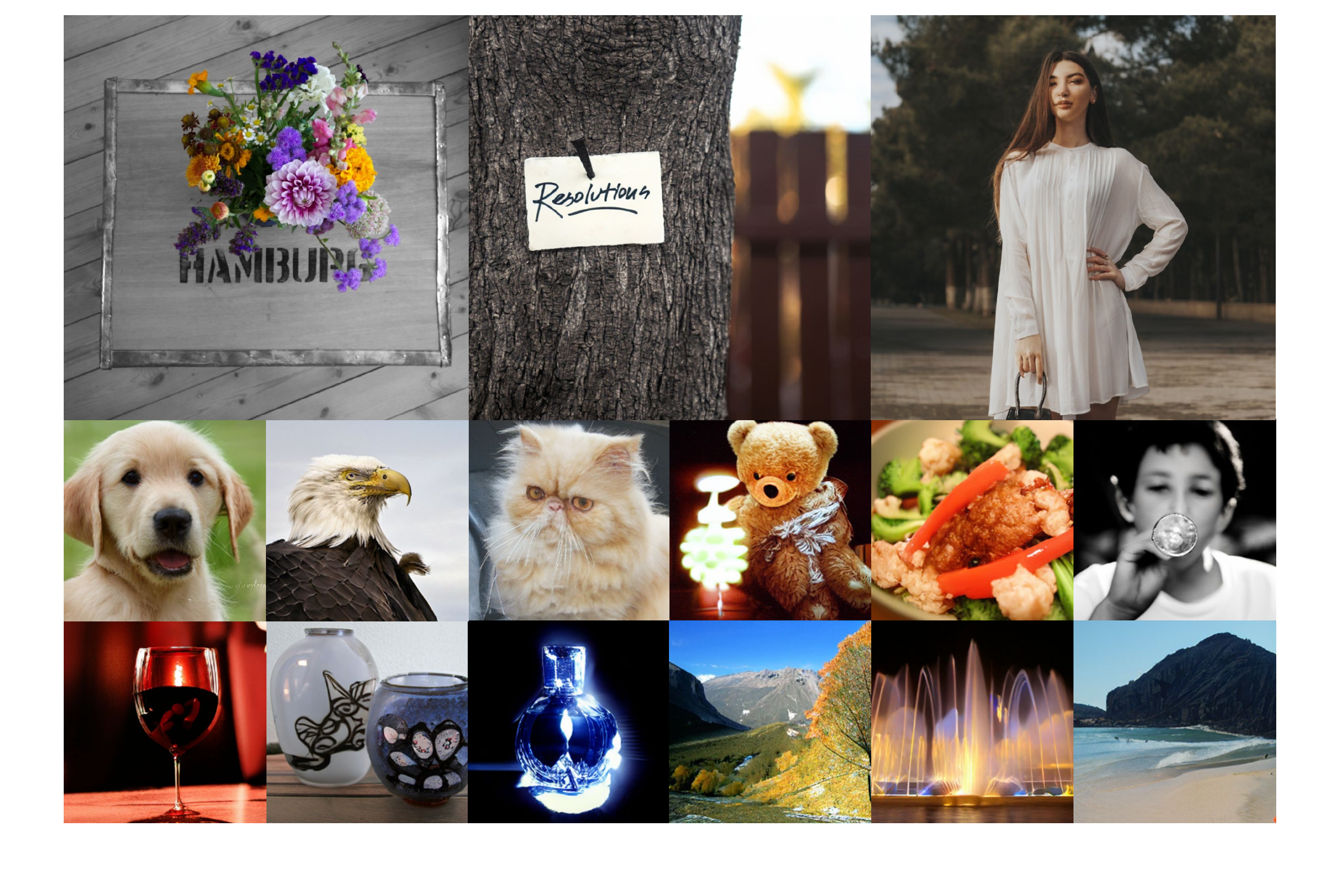}
\vspace{-2.5mm}
\captionof{figure}{\textbf{Reconstruction and generation samples of IBQ.} We show 1024 $\times$ 1024 reconstructed samples (top) and 256 $\times$ 256 generated samples (middle and bottom).}
\vspace{5mm}
\label{fig:first_page}
}]

\input{sec/0_abstract}    
\input{sec/1_intro}
\input{sec/2_related_work}

\input{sec/3_method}
\input{sec/4_experiment}
\input{sec/5_conclusion}
\input{sec/X_suppl}

{
    \small
    \bibliographystyle{ieeenat_fullname}
    \bibliography{main}
}

\end{document}

%% file: preamble.tex
\usepackage{amsmath}
\DeclareMathOperator*{\argmin}{arg\,min}
\usepackage[misc]{ifsym}
\usepackage{multirow}

%% file: sec/0_abstract.tex
\begin{abstract}
\label{sec:intro}
\begin{figure*}
    \centering
    \includegraphics[width=1.0\linewidth]{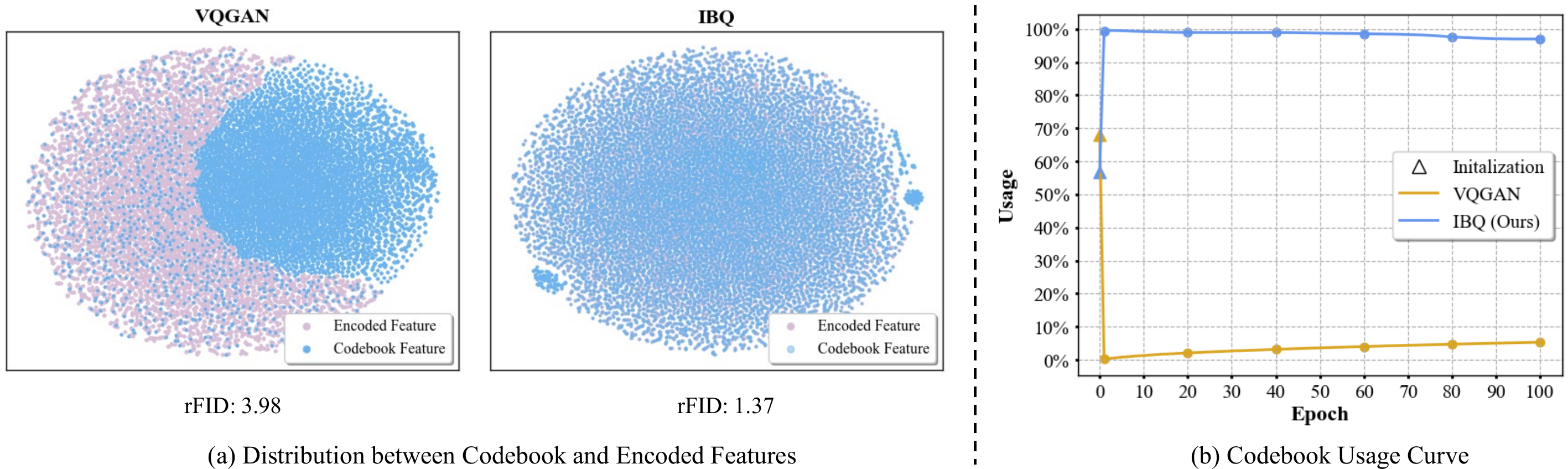}
    \vspace{-7mm}
    \caption{\textbf{Effects of Distribution Gap on Codebook Usage.} (a) T-SNE of the codebook (16,384 codebook size and 256 dimension) and sampled encoder features. (b) Codebook usage curve. The partial-update strategy adopted by VQGAN broadens the distribution gap between encoder features and non-activated codes, while those of IBQ based on all-codes updating are evenly mixed, maintaining a high codebook usage ($\sim$96\%) throughout the training. Fully leveraging the codebook significantly improves the reconstruction quality.}
    \label{fig:dis_gap}
    \vspace{-4mm}
\end{figure*}
Existing vector quantization (VQ) methods struggle with scalability, largely attributed to the instability of the codebook that undergoes partial updates during training. The codebook is prone to collapse as utilization decreases, due to the progressively widening distribution gap between non-activated codes and visual features. To solve the problem, we propose {I}ndex {B}ackpropagation {Q}uantization (IBQ), a new VQ method for the joint optimization of all codebook embeddings and the visual encoder. 
Applying a straight-through estimator on the one-hot categorical distribution between the encoded feature and codebook, all codes are differentiable and maintain a consistent latent space with the visual encoder. IBQ enables scalable training of visual tokenizers and, for the first time, achieves a large-scale codebook ($2^{18}$) with high dimension ($256$) and high utilization. Experiments on the standard ImageNet benchmark demonstrate the scalability and superiority of IBQ, achieving competitive results on reconstruction and the application of autoregressive visual generation. The code and models are available at \url{https://github.com/TencentARC/SEED-Voken}.
\end{abstract}

%% file: sec/1_intro.tex
\section{Introduction}
Discrete tokenizer plays a pivotal role in processing complex data across various modalities, such as text~\cite{gpt3, openai2023gpt4, llama2}, images~\cite{vqvae, vqgan, beit}, and audio~\cite{vq-wav2vec,audiolm}. By transforming raw data into discrete tokens, models can effectively handle diverse data types within a unified framework~\cite{attention, gpt3}, simplifying the integration of multimodal information and facilitating native large multimodal models~\cite{chameleon, emu3}.

In the image domain, pioneering works like VQGAN~\cite{vqgan} employ vector quantization (VQ) to learn visual tokenizers, enabling effective data compression and reconstruction.
VQ tokenizers are deemed as the key component in applications such as autoregressive image generation~\cite{vqvae, vqgan, llamagen, var} and representation learning~\cite{beit, mage}. However, a notable challenge with VQ-based methods is the information loss during quantization, leading to inferior reconstruction performance compared to continuous representation models like Variational Autoencoders (VAEs)~\cite{vae, ldm}. 
Intuitively, scaling visual tokenizers by increasing the codebook size and embedding dimension could help mitigate the information loss associated with discrete tokens, thereby bridging the gap between discrete and continuous representations. However, it is noteworthy that current visual tokenizers~\cite{vit-vqgan,llamagen} have not demonstrated such scaling properties.

Empirical research~\cite{vqgan, llamagen} has revealed that current VQ methods struggle with scalability due to the inherent tendency of the codebook to collapse. This arises because these methods only optimize a limited number of the selected codes during each backpropagation. Such a widely-adopted partial update strategy gradually broadens the distribution gap between non-activated codes and the visual encoder's representation space, making the non-activated codes further less likely to be selected. As shown in \cref{fig:dis_gap}, VQGAN~\cite{vqgan} almost fails when scaling both codebook size (\ie, $16,384$) and embedding dimension (\ie, $256$) simultaneously. Only a small amount of codes share the same distribution with the visual encoder, and the codebook usage degrades from $68\%$ to $0.002\%$ after training one epoch.

To tackle the challenge, we introduce a new VQ method, namely, Index Backpropagation Quantization (\textbf{IBQ}). It globally updates the entire codebook in each backward process to ensure consistency with the distribution of the visual encoder. In such a way, all codes have the same probability of being selected, resulting in a high utilization of the codebook throughout the training process. Specifically, rather than directly applying the straight-through estimator~\cite{ste} to the selected codes, we employ this reparameterization approach on the categorical distribution between visual features and all codebook embeddings, thereby rendering all codes differentiable. As shown in \cref{fig:dis_gap}, the sampled visual features and the codebook embeddings from IBQ are evenly mixed. IBQ keeps a high codebook usage ($\sim96\%$) throughout the training process. Fully utilizing the codebook effectively enhances the representation capacity, as demonstrated by the superior reconstruction of IBQ ($1.37$ rFID) compared to VQGAN ($3.98$ rFID).

We conduct a comprehensive study on the scaling behavior of IBQ tokenizers along three axes: codebook size, code dimension, and model size. We observe significant gains in reconstruction quality or codebook usage when scaling up tokenizers. To our knowledge, IBQ is the pioneering work to train an extremely large codebook (\ie, $262,144$) with a relatively large code dimension (\ie, $256$). This achievement leads to state-of-the-art reconstruction quality, reaching an rFID of $1.00$. We further demonstrate the effectiveness of IBQ tokenizers in autoregressive image generation by integrating them with vanilla transformers of varying scales, ranging from 300M to 2.1B parameters, achieving competitive performance. Although IBQ is also compatible with other advanced autoregressive models, the paper does not focus on them, leaving them for future research.

In summary, our contributions are threefold:
\begin{itemize}
    \item We propose a simple yet effective vector quantization method, dubbed Index Backpropagation Quantization (IBQ), for training scalable visual tokenizers.
    \item We study the scaling properties of IBQ by increasing codebook size, code dimension, and model size. IBQ for the first time trains a super large codebook ($2^{18}$) with a large dimension ($256$) and high usage, achieving state-of-the-art reconstruction performance.
    \item We validate the effectiveness of IBQ tokenizers in visual generation by equipping them with vanilla autoregressive models ranging from 300M to 2.1B, remarkably outperforming competing methods, \eg, LlamaGen \cite{llamagen}, and Open-MAGVIT2 \cite{open-magvit2}.
\end{itemize}

%% file: sec/2_related_work.tex
\section{Related Work}
\label{sec:related_work}

\subsection{Vector Quantization}
\label{sec:related_vq}
At the core of visual tokenizers is vector quantization, which maps the visual signals into discrete tokens. VQ-VAE~\cite{vqvae} proposes an encoder-quantizer-decoder structure with a learnable codebook as the discrete representation space. VQ-VAE2~\cite{vqvae2} introduces multi-scale hierarchical VQ-VAE to enhance local features. VQGAN~\cite{vqgan} further uses adversarial loss and perceptual loss for good perceptual quality. RQ-VAE~\cite{rqvae} and DQ-VAE~\cite{dqvae} improve VQGAN by residual quantization and dynamic quantization, respectively. To improve codebook utilization for large-size codebooks, some works try to decrease code dimension~\cite{vit-vqgan, llamagen}. Following this observation, MAGVIT-v2~\cite{magvit2} reduces the code dimension to zero, and expands the codebook size to $2^{18}$ with Lookup-Free Quantization. Instead of joint optimization of the model and codebook, VQGAN-LC~\cite{vqganlc} extends the codebook size to 100,000 using a frozen codebook with a trainable projector. However, it introduces a bottleneck that constrains the tokenizer capacity.

 Existing VQ methods suffer from codebook collapse when scaling up tokenizers and typically use small-size codebooks or low-dimensional code embeddings, limiting the representational capacity. In contrast, our proposed IBQ shows consistent improvements when scaling up codebook size, code dimension and model size.

\subsection{Tokenized Visual Generation}
\label{sec:related_vg}
Tokenizers map continuous visual signals into a discrete token sequence. For subsequent visual generation, there are two approaches, including non-autoregressive (NAR) and autoregressive (AR) generation. 
NAR~\cite{maskgit, magvit2} usually adopts BERT-style transformers to predict masked tokens. For inference, these methods generate all tokens of an image simultaneously, and iteratively refine the generated images conditioned on the previous generation. In contrast, AR models perform next-token prediction in a raster-scan manner. VQGAN~\cite{vqgan} adopts GPT2-medium architecture, while LlamaGen~\cite{llamagen} employs Llama~\cite{llama2} for scalable image generation. VAR~\cite{var} extends ``next-token prediction'' to ``next-scale prediction'' and introduces adaptive normalization (AdaLN~\cite{dit}) to improve generation quality. Open-MAGVIT2~\cite{open-magvit2} proposes asymmetric token factorization for super-large codebook learning. 

In this paper, we adopt vanilla autoregressive models to validate the effectiveness of IBQ tokenizers in visual generation, excluding masked modeling or multi-scale structures for simplicity. Notably, IBQ is compatible with advanced generative models, which can further unlock the tokenizer potential. We leave this exploration for future work.

\input{table/algrithom}

\begin{figure*}
  \centering
  \includegraphics[width=0.9\linewidth]{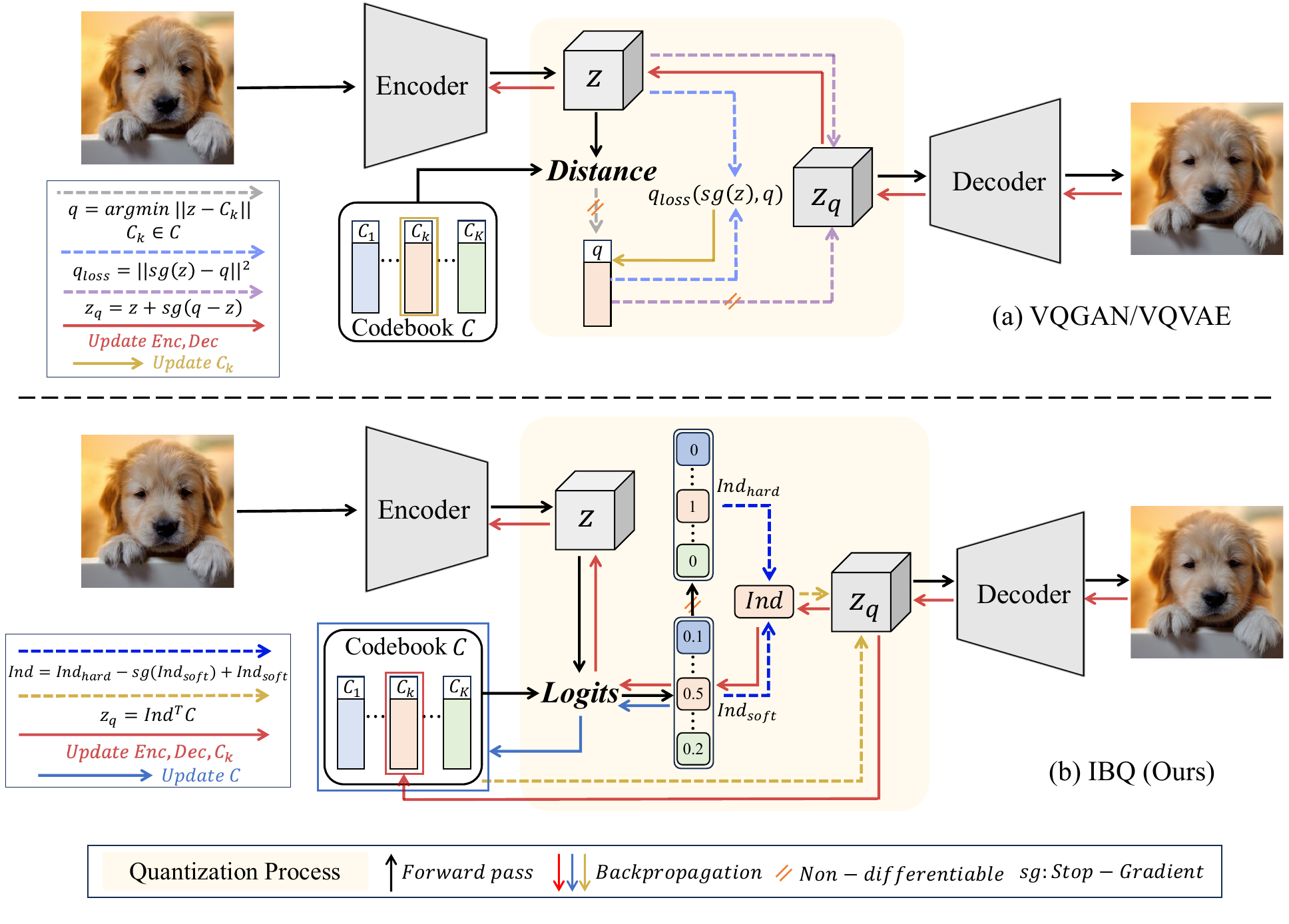}
  \vspace{-2mm}
  
  \caption{\textbf{Gradient flow of different VQ methods.} VQGAN/VQVAE only update the selected codes in each backward process. IBQ updates all codes simultaneously by transferring the gradients of soft one-hot categorical distribution to hard one-hot index.}
  \label{fig:method_comparison}
  \vspace{-5mm}
\end{figure*}

%% file: table/algrithom.tex
\begin{algorithm}[t]
\caption{Pseudocode of IBQ in a PyTorch-like style}
\algcomment{\fontsize{7.2pt}{0em}\selectfont \texttt{mm}: matrix multiplication; \texttt{onehot}: transfer index into one-hot vector.
\vspace{-5mm}
}
\definecolor{codeblue}{rgb}{0.25,0.5,0.5}
\definecolor{codekw}{rgb}{0.85, 0.18, 0.50}

\begin{lstlisting}[language=python]
def IBQ(z, codebook):
    '''
    z: visual feature map (B * h * w, D)
        B: batch size
        h: height of feature map
        w: width of feature map
        D: feature dimension
    codebook: (K, D)
        K: codebook size
        D: code dimension
    '''
    logits = mm(z, codebook.T) # (B * h * w, K)
    Ind_soft = softmax(logits, dim=1) # (B * h * w, K)
    _, indices = soft_one_hot.max(dim=1)
    Ind_hard = onehot(indices) # (B * h * w, K)
    Ind = Ind_hard - Ind_soft.detach() + Ind_soft
    z_q = mm(Ind, codebook) # (B * h * w, D)
    return z_q
\end{lstlisting}
\vspace{-2mm}
\label{method:alg_ibq}
\end{algorithm}

%% file: sec/3_method.tex
\section{Method}
\label{sec:method}

\subsection{Preliminary: Vector Quantization}
\label{sec:method_pre}
Vector quantization (VQ) maps continuous visual signals into discrete tokens with a fixed-size codebook $\mathcal{C} \in \mathbb{R}^{K \times D}$, where $K$ is the codebook size and $D$ is the code dimension. Given an image $\mathcal{I} \in \mathbb{R}^{H \times W \times 3}$, VQ first utilizes an encoder to project the image into the feature map $\mathcal{Z} \in \mathbb{R}^{h \times w \times D}$, where $h = H / p$, $w = W / p$, and $p$ is the downsample ratio. The feature map is then quantized into $\mathcal{Q} \in \mathbb{R}^{h \times w \times D}$ discrete representations using the codebook. Finally, the decoder reconstructs the image given the quantized features. 

Previous methods quantize each visual feature $z\in \mathbb{R}^D$ by selecting the nearest code from the codebook based on Euclidean distance~\cite{vqvae, vqgan}. 
Since the $\argmin$ operation in quantization is non-differentiable, they apply the straight-through estimator on the selected codes to copy the gradients from the decoder to the encoder, to optimize the encoder and decoder simultaneously. The quantization process can be formulated as:
\begin{align}
    q &= \argmin_{\mathcal{C}_k \in \mathcal{C}}||z - \mathcal{C}_k|| \in \mathbb{R}^D, \\
    z_q &= z + \text{sg}[q - z],
\end{align}
where sg$[\cdot]$ is stop-gradient operation.

This partial updating strategy (\ie, only selected codes are optimized) adopted by these methods progressively widens the distribution gap between visual features and non-activated codes. It incurs the instability during training due to the codebook collapse, which hampers the scalability of the visual tokenizer.

\subsection{Index Backpropagation Quantization}
\label{sec:method_ibq}

\noindent \textbf{Quantization.} To ensure the consistent distribution between the codebook and encoded features through the training, we introduce an all-codes updating method, Index Backpropagation Quantization (IBQ). The core idea of IBQ is to pass gradients to all codebook embeddings, rather than the selected ones only. \cref{method:alg_ibq} provides the pseudo-code of IBQ.

Specifically, we first perform dot product between the given visual feature $z$ and all code embeddings as logits and get probabilities (soft one-hot) by softmax function.
\begin{align}
    \text{logits} &= [z^T\mathcal{C}_1, z^T\mathcal{C}_2, \cdots, z^T\mathcal{C}_K]^T \in \mathbb{R}^K, \\
    \text{Ind}_{\text{soft}} &= \text{softmax}(\text{logits}), \\
    \text{Ind}_{\text{hard}} &= \text{One-Hot}(\text{argmax}(\text{Ind}_{\text{soft}})).
\end{align}
We then copy the gradients from soft one-hot categorical distribution to hard one-hot index:
\begin{align}
    \text{Ind} = \text{Ind}_{\text{hard}} - \text{sg}[\text{Ind}_{\text{soft}}] + \text{Ind}_{\text{soft}}.
\end{align}
Given the index, the quantized feature can be computed as:
\begin{align}
    z_q= \text{Ind}^T\mathcal{C}.
\end{align}
In this way, we can pass the gradients to all codes of the codebook via index. By Index Backpropagation Quantization, the distribution of the whole codebook and encoded features remains consistent throughout completed training, thus gaining a high codebook utilization.

\noindent \textbf{Training Losses.} Similar to VQGAN~\cite{vqgan}, the tokenizer is optimized with a combination of losses:
\begin{align}
    \mathcal{L} &= \mathcal{L}_R + \mathcal{L}_Q + \mathcal{L}_P + \mathcal{L}_G + \mathcal{L}_E,
\end{align}
where $\mathcal{L}_R$ is reconstruction loss of image pixels, $\mathcal{L}_Q$ is quantization loss between the selected code embeddings and encoded features, $\mathcal{L}_P$ is perceptual loss from LPIPS~\cite{lpips}, $\mathcal{L}_G$ is adversarial loss with PatchGAN discriminator~\cite{patchgan} to enhance the image quality, and $\mathcal{L}_E$ is entropy penalty to encourage codebook utilization~\cite{magvit2}.

To better explain how IBQ keeps the consistent distribution between the encoder features and the whole codebook, we provide a gradient analysis. Considering the quantization loss $\mathcal{L}_Q = ||z-z_q||^2$, 
\begin{align}
\frac{\partial \mathcal{L}_Q}{\partial \mathcal{C}_k} &= -2\text{Ind}_k(z-z_q) = -2p_k(z-z_q), \\
\ p_k &= \frac{\text{exp}(z^T\mathcal{C}_k)}{\sum_{j=1}^{K}\text{exp}(z^T\mathcal{C}_j)}.
\end{align}
The softmax probabilities $p_k$ ensure that each $\mathcal{C}_k$ is updated based on its similarity to the encoder feature $z$, and $z-z_q$ shifts $\mathcal{C}_k$ toward dominant regions of the feature distribution $P_Z(z)$. Random batch sampling covers the whole encoder latent space, gradually aligning the entire codebook $\mathcal{C}$ with the distribution $P_Z(z)$ of encoder features over time.

We further introduce double quantization loss, to force the selected code embeddings and given encoded visual features towards each other.
\vspace{-2mm}
\begin{align}
    {z_q}' = &\text{Ind}_{\text{hard}}^T \mathcal{C}, \\
    \begin{split}
    \mathcal{L}_Q = &||z_q - z||^2 + \\
                  & ||\text{sg}[z] - {z_q}'||^2 + \beta ||z - \text{sg}[{z_q}']||^2.
    \end{split}
\end{align}

\noindent \textbf{Discussion with other VQ methods.}
As shown in \cref{fig:method_comparison}, existing VQ methods (\eg, VQ-VAE~\cite{vqvae} and VQGAN~\cite{vqgan}) update only a few selected codes within each backward process, progressively widening the gap between non-activated codes and encoded features, which leads to codebook collapse. This issue worsens as code dimension and codebook size increase. Instead of applying the straight-through estimator~\cite{ste} on the selected codes, we employ it on the categorical distribution between visual features and all codebook embeddings to enable gradients backward to all codes. This promotes distribution consistency between the codebook and encoded features throughout training, allowing IBQ to scale up to extremely large codebook size with high code dimension and utilization.

\vspace{-1mm}
\subsection{Vanilla Autoregressive Transformer}
\label{sec:method_ar}
After tokenization,  the visual feature is quantized into discrete representations which are subsequently flattened in a raster-scan manner for visual generation. Given the discrete token index sequence $\mathcal{X} = \{x_i\}_{i=1}^T$, where $T = h' \times w'$, we employ an autoregressive transformer to model the sequence dependency through next-token prediction. Specifically, the optimization process is to maximize the log-likelihood:
\begin{equation}
p(x_1, \cdots, x_{T}|c) = \prod_{t=1}^{T}p(x_t|x_1, \cdots, x_{t-1}, c),
\end{equation}
where $c$ is the condition such as class label. 

Note that, since our focus is on the visual tokenizer, we adopt the vanilla architecture of autoregressive transformers akin to Llama~\cite{llama2} with AdaLN~\cite{dit} for visual generation. More details can be deferred to the supplementary. 

\input{table/reconstruction}
\input{table/ablation_scale}

\begin{figure*}
  \centering
  \includegraphics[width=0.9\linewidth]{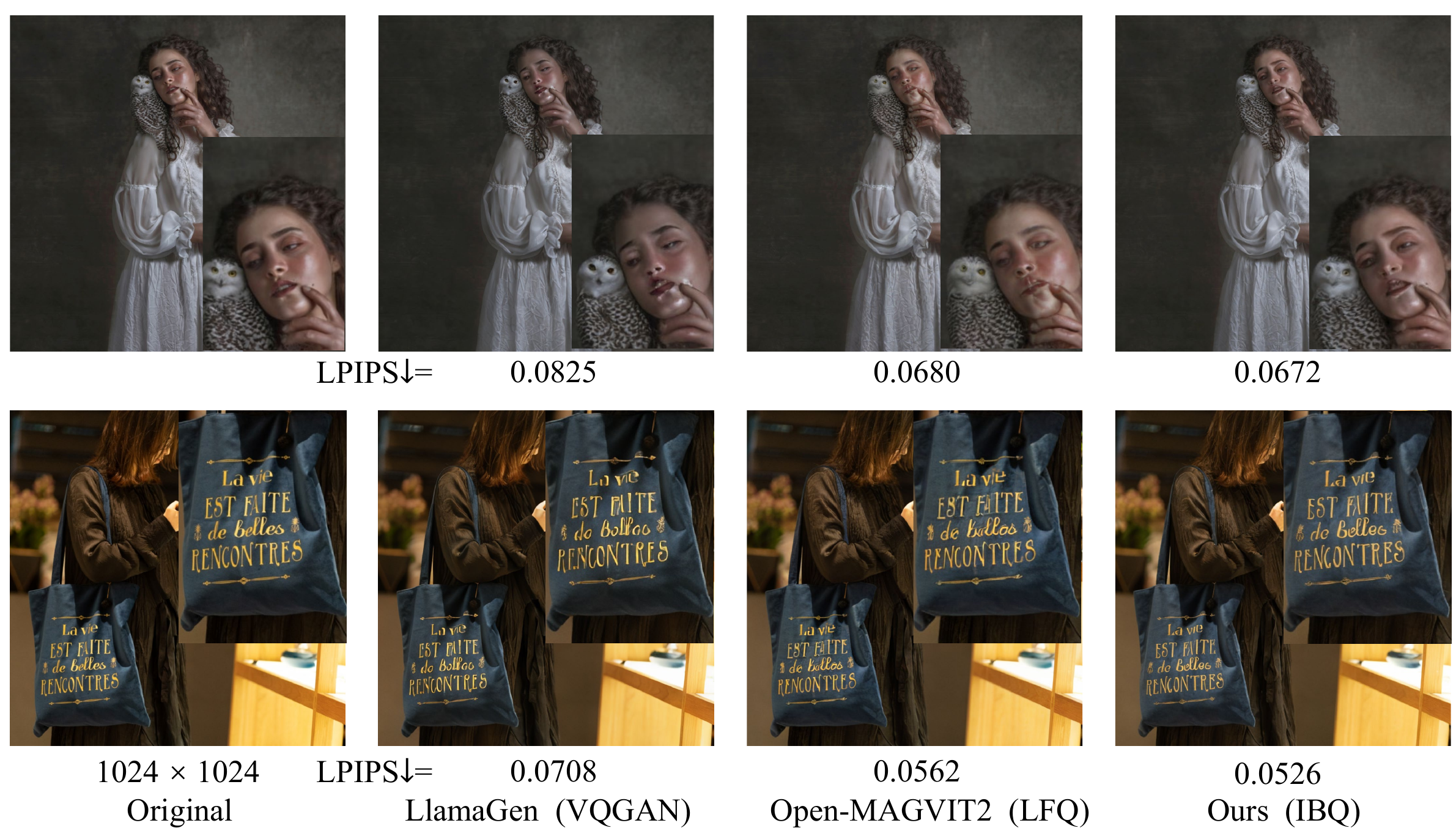}
  \vspace{-3mm}
  \caption{\textbf{Qualitative Reconstruction Comparison.} We compare IBQ with LlamaGen and Open-MAGVIT2 tokenizer.}
  \label{fig:construction_comparison}
  \vspace{-5mm}
\end{figure*}

%% file: table/reconstruction.tex
\begin{table*}[t]
    \centering
    \footnotesize
    \setlength{\tabcolsep}{4pt}
    \renewcommand\arraystretch{1.1}
    \resizebox{0.96\linewidth}{!}{
        \begin{tabular}{lccccccccc}
        \toprule
        \multirow{2}{*}{\textbf{Method}} & \textbf{Token} & \multirow{2}{*}{\textbf{Tokens}} & \multirow{2}{*}{\textbf{Ratio}} & \textbf{Train} & \textbf{Codebook} & \textbf{Codebook} & \multirow{2}{*}{\textbf{rFID}$\downarrow$} & \multirow{2}{*}{\textbf{LPIPS}$\downarrow$} & \textbf{Codebook} \\ 
        & \textbf{Type} &  & & \textbf{Resolution} & \textbf{Size} & \textbf{Dim}& & & \textbf{Usage}$\uparrow$ \\
        \midrule
        VQGAN~\cite{vqgan} & 2D & 16 $\times$ 16 & 16 & 256 $\times$ 256 & 1,024  & 256 & 7.94 & $-$ & 44\% \\
        VQGAN~\cite{vqgan} & 2D & 16 $\times$ 16 & 16 & 256 $\times$ 256 & 16,384  & 256 & 4.98 & 0.2843 & 5.9\% \\
        VQGAN$^{*}$~\cite{vqgan} & 2D & 16 $\times$ 16 & 16 & 256 $\times$ 256 & 16,384  & 256 & 3.98 & 0.2873 & 5.3\% \\
        SD-VQGAN~\cite{ldm} & 2D & 16 $\times$ 16 & 16 & 256 $\times$ 256 & 16,384 & 8 & 5.15 & $-$ & $-$ \\
        MaskGIT~\cite{maskgit} & 2D & 16 $\times$ 16 & 16 & 256 $\times$ 256 & 1,024 & 256 & 2.28 & $-$ & $-$ \\
        LlamaGen~\cite{llamagen} & 2D & 16 $\times$ 16 & 16 & 256 $\times$ 256 & 16,384 & 256 & 9.21 & $-$ & 0.29$\%$ \\
        LlamaGen~\cite{llamagen} & 2D & 16 $\times$ 16 & 16 & 256 $\times$ 256 & 16,384 & 8 & 2.19 & 0.2281 & 97$\%$ \\
        VQGAN-LC~\cite{vqganlc} & 2D & 16 $\times$ 16 & 16 & 256 $\times$ 256 & 16,384 & 8 & 3.01 & 0.2358 & 99$\%$ \\
        VQGAN-LC~\cite{vqganlc} & 2D & 16 $\times$ 16 & 16 & 256 $\times$ 256 & 100,000 & 8 & 2.62 & 0.2212 & 99$\%$ \\
        Open-MAGVIT2~\cite{open-magvit2} & 2D & 16 $\times$ 16 & 16 & 256 $\times$ 256 & 16,384 & 0 & 1.58 & 0.2261 & 100\% \\ 
        Open-MAGVIT2~\cite{open-magvit2} & 2D & 16 $\times$ 16 & 16 &256 $\times$ 256 & 262,144 & 0 & 1.17 & 0.2038 & 100\% \\
        \textbf{IBQ (Ours)} & 2D & 16 $\times$ 16 & 16 & 256 $\times$ 256 & 16,384 & 256 & $1.37$ & 0.2235 & 96$\%$ \\
        \textbf{IBQ (Ours)} & 2D & 16 $\times$ 16 & 16 & 256 $\times$ 256 & 262,144 & 256 & \textbf{1.00} & \textbf{0.2030} & 84\% \\
        \hline
        Titok-L~\citep{titok} & 1D & 32 & $-$ & $256 \times 256$ & 4,096 & 16 & 2.21 & $-$ & $-$ \\
        Titok-B~\citep{titok} & 1D & 64 & $-$ & $256 \times 256$ & 4,096 & 16& 1.70 & $-$ & $-$ \\
        Titok-S~\citep{titok} & 1D & 128 & $-$ & $256 \times 256$ & 4,096 & 16 & 1.71 & $-$ & $-$ \\
    \bottomrule
    \end{tabular}}
    \vspace{-2mm}
    \caption{\textbf{Reconstruction performance of different tokenizers on $\boldsymbol{256 \times 256}$ ImageNet 50k validation set.} $^{*}$ reproduced VQGAN.}
    \label{tab:recon1}
\end{table*}

%% file: table/ablation_scale.tex
\begin{table*}[t]
\vspace{-6pt}
\begin{minipage}[c]{\textwidth}
    \begin{minipage}{0.4\textwidth}
    \makeatletter\def\@captype{table}
    \centering
    \footnotesize
    \setlength{\tabcolsep}{2.5pt}
    \begin{tabular}{cccccc}
    \toprule
      IBQ & Double Quant. & Deeper Model & rFID$\downarrow$ & LPIPS$\downarrow$ & Usage$\uparrow$ \\
      \midrule
        & & & $3.98$ & $0.2873$ & $5.3\%$ \\
        $\checkmark$& & & 1.67 & 0.2340 & 98\% \\
        $\checkmark$ & $\checkmark$ & & 1.55 & 0.2311 & 97\%\\
        $\checkmark$ & $\checkmark$ & $\checkmark$ & 1.37 & 0.2235 & 96\% \\
        \bottomrule
    \end{tabular}
    \vspace{-3mm}
    \caption{\textbf{Effectiveness of designed modules.}}
    \label{tab:module_ablation}
    \end{minipage}
    \hspace{0.02\textwidth}
    \begin{minipage}{0.28\textwidth}
    \makeatletter\def\@captype{table}
    \centering
    \footnotesize
    \setlength{\tabcolsep}{2.5pt}
    \begin{tabular}{c|ccc}
    \toprule
    Codebook Size & rFID$\downarrow$ & LPIPS$\downarrow$ & Usage$\uparrow$ \\
    \midrule
    1,024 & 2.24 & 0.2580 & 99\% \\
    8,192 & 1.87 & 0.2437 & 98\% \\
    16,384 & 1.37 & 0.2235 & 96\% \\
    262,122 & 1.00 & 0.2030 & 84\% \\
    \bottomrule
    \end{tabular}
    \vspace{-3mm}
    \caption{\textbf{Impact of codebook size.}}
    \label{tab:codebook_size}
    \end{minipage}
    \hspace{0.01\textwidth}
    \begin{minipage}{0.28\textwidth}
    \makeatletter\def\@captype{table}
    \centering
    \footnotesize
    \setlength{\tabcolsep}{2.5pt}
    \begin{tabular}{c|ccc}
    \toprule
    Codebook Dim & rFID$\downarrow$ & LPIPS$\downarrow$ & Usage$\uparrow$ \\
    \midrule
    32   &  2.04 & 0.2408 & 92\% \\ 
    64   &  1.39 & 0.2281 & 69\% \\
    128  &  1.38 & 0.2255 & 77\% \\
    256  &  1.37 & 0.2235 & 96\% \\
    \bottomrule
    \end{tabular}
    \vspace{-3mm}
    \caption{\textbf{Effect of codebook dim.}}
    \label{tab:codebook_dim}
    \end{minipage}
    \begin{minipage}{0.4\textwidth}
    \makeatletter\def\@captype{table}
    \centering
    \footnotesize
    \setlength{\tabcolsep}{3.0pt}
    \begin{tabular}{c|ccc}
    \toprule
    Num Resblock &  rFID$\downarrow$ & LPIPS$\downarrow$ & Usage$\uparrow$\\
    \midrule
    1   & 1.80 & 0.2377 & 99\% \\
    2   & 1.55 & 0.2311 & 97\% \\
    4   & 1.37 & 0.2235 & 96\% \\
    \bottomrule
    \end{tabular}
    \vspace{-2.5mm}
    \caption{\textbf{Benefit of larger model size.}}
    \label{tab:model_size}
    \end{minipage}
    \vspace{-3mm}
    \begin{minipage}{0.55\textwidth}
    \makeatletter\def\@captype{table}
    \centering
    \footnotesize
    \setlength{\tabcolsep}{3.0pt}
    \begin{tabular}{cccccc|cc}
    \toprule
    \multirow{2}{*}{Method} & Codebook & Codebook & Transformer & \multirow{2}{*}{rFID$\downarrow$} & \multirow{2}{*}{LPIPS$\downarrow$} & \multirow{2}{*}{gFID$\downarrow$} & \multirow{2}{*}{IS$\uparrow$} \\
    & size & dim &scale & & & &\\
    \midrule
    LFQ & 16,384 & 0 &342M & 1.58 & 0.2261 & 3.40 &228.03 \\
    IBQ & 16,384 & 256 &342M & 1.37 & 0.2235 & 2.88 & 254.73 \\
    \bottomrule
    \end{tabular}
    \vspace{-2.5mm}
    \caption{\textbf{Performance comparison with LFQ.}}
    \label{tab:lfq_compare}
    \end{minipage}
    \vspace{-3mm}
\end{minipage}
\end{table*}

%% file: sec/4_experiment.tex
\section{Experiment}
\label{sec:experiment}
\begin{figure}[t]
    \centering
    \includegraphics[width=1.0\linewidth]{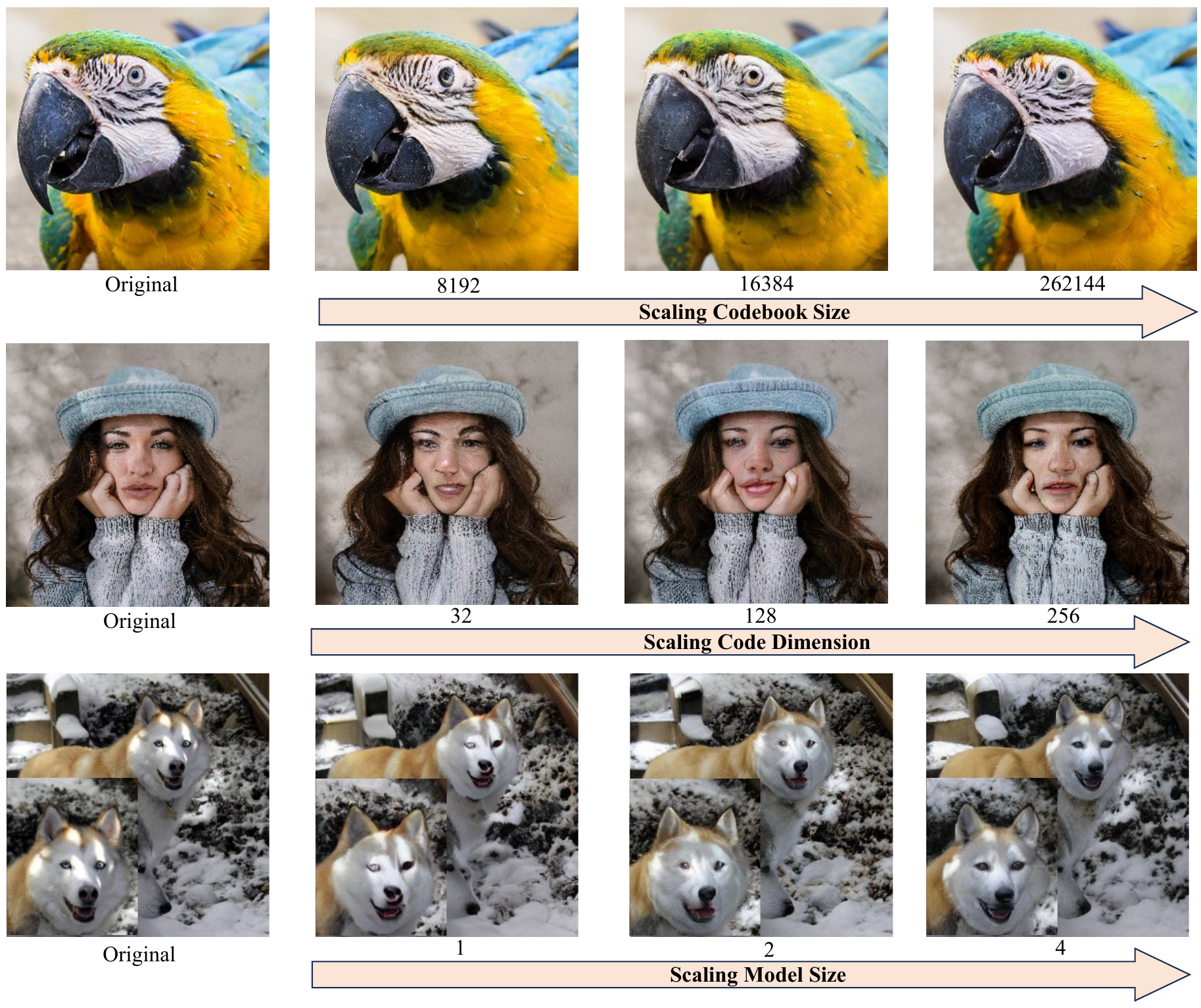}
    \vspace{-8mm}
    \caption{\textbf{Scaling up visual tokenizers (\eg, codebook size, code dimension and model size) improves visual soundness of reconstruction.}}
    \label{fig:scaling_visualization}
    \vspace{-8mm}
\end{figure}

\input{table/generation}
\subsection{Datasets and Metrics}
Both the visual tokenizers and autoregressive transformers are trained on $256 \times 256$ ImageNet~\cite{imagenet}. For reconstruction, we measure reconstruction-FID (rFID~\cite{fid}), codebook utilization, and LPIPS~\cite{lpips} on the ImageNet 50k validation set. For generation, we evaluate image quality using generation FID (gFID), Inception Score~\citep{is}, and Precision/Recall~\cite{precision_recall}, following ADM evaluator~\cite{adm}.

\subsection{Implementations Details}
\paragraph{Visual Reconstruction Setup.}
We adopt the same model architecture proposed in VQGAN~\cite{vqgan}. The visual tokenizer is trained with the following settings: an initial $1e-4$ learning rate with $0.01$ multi-step decay mechanism, an Adam Optimizer~\cite{adam} with $\beta_1 = 0.5$, $\beta_2=0.9$, a total $256$ batch size with $330$ epochs, a combination of reconstruction, GAN~\cite{patchgan}, perceptual~\cite{lpips}, commitment~\cite{vqgan}, entropy~\cite{magvit2}, double quantization losses, and LeCAM regularization~\cite{lecam} for training stability. Unless otherwise specified, we use a codebook size of 16,384, a code dimension of 256, and 4 ResBlocks as our default tokenizer setting.

\vspace{-4mm}

\paragraph{Visual Generation Setup.}
We use vanilla Autoregressive models ranging from 300M to 2.1B to validate the effectiveness of IBQ tokenizers in visual generation, adopting a Llama-based architecture with RoPE~\cite{rope}, SwiGLU~\cite{glu}, and RMSNorm~\cite{rmsnorm}. AdaLN~\cite{dit} is also incorporated for improved visual synthesis quality. The class embedding serves as both the start token and AdaLN condition. IBQ with width $w$, depth $d$ and head $h$ follows the scaling rules proposed in~\cite{llamagen, var}, where $w=64d, h=d$. All models are trained with similar settings: a base learning rate of $1e-4$ per $256$ batch size, an AdamW optimizer~\cite{adamw} with $\beta_{1} = 0.9$, $\beta_2 = 0.95$, weight decay = $5e-2$, a total $768$ batch size and 300 $\sim$ 450 training epochs corresponding the model size, gradient clipping of $1.0$, and $0.1$ dropout rate for input embedding, FFN module and conditional embedding.

\subsection{Main Results}

\paragraph{Visual Reconstruction.} 
\cref{tab:recon1} shows the quantitative reconstruction comparison between IBQ and prevalent visual tokenizers. Existing VQ methods show a significant drop in codebook usage when scaling codebook size (\eg, VQGAN~\cite{vqgan} has a 44\% usage for 1024 codebook size, while a 5.9\% usage for 16,384 codebook size.), and code dimension (\eg, LlamaGen~\cite{llamagen} has a 97\% usage for 8-dimension codes, while a 0.29\% usage for 256-dimension codes.) Therefore, the actual representational capacity is limited by the codebook collapse. 

In contrast, IBQ's joint optimization of codebook embeddings and the visual encoder maintains distribution consistency, enabling stable training of large-scale codebook with high utilization. Specifically, IBQ with 16,384 codebook size and 256 code dimension achieves 1.37 rFID, outperforming other VQ methods at the same downsampling rate and codebook size. Increasing the codebook size to 262,144, IBQ achieves state-of-the-art reconstruction with 1.00 rFID, surpassing Open-MAGVIT2~\cite{open-magvit2}. A qualitative comparison in \cref{fig:construction_comparison} shows IBQ's superior visual quality in complex scenarios such as faces and characters. Note that, we observe that incorporating additional facial data yields consistent improvements (see supplementary materials).

\vspace{-4mm}

\paragraph{Visual Generation.} 
In \cref{tab:gen1}, we compare IBQ with other generative models, including Diffusion models, AR models, and variants of AR models (VAR~\cite{var} and MAR~\cite{mar}) on class-conditional image generation. With powerful IBQ tokenizers, our models show consistent improvements when scaling up the model size (from 300M to 2.1B), and outperform all previous vanilla autoregressive models at different scales of model size. Moreover, IBQ outperforms the diffusion-based model DiT~\cite{dit}, and achieves comparable results with the variants of AR models. These AR model variants focus on the architecture designs of transformers in the second stage, while our work is devoted to better visual tokenizers in the first stage. Therefore, we believe that with our stronger tokenizers, the AR models and their variants can be boosted further.

\subsection{Scaling Up IBQ Tokenizers}
Existing VQ methods struggle to scale up due to the codebook collapse. For example, LlamaGen~\cite{llamagen} sees a significant drop in usage and rFID when increasing the code dimension from 8 to 256 (97\% $\rightarrow$ 0.29\%, 2.19 rFID $\rightarrow$ 9.21 rFID), as shown in \cref{tab:recon1}. This is due to their partial updates during training, which progressively widens the distribution gap between non-activated codes and encoded features.

\noindent \textbf{Scaling Up Tokenizers Improves Reconstruction.} IBQ tokenizers show promising scaling capacity for reconstruction in three aspects: \textbf{1) Codebook Size}: As shown in \cref{tab:codebook_size}, reconstruction quality improves significantly as the codebook size increases from 1,024 to 16,384, with high utilization and consistent visual soundness even at 262,144 codes. \textbf{2) Code Dimension}: interestingly, we observe a notable increase in codebook usage when scaling code dimension in \cref{tab:codebook_dim}. We assume that low-dimensional codes are less discriminative and tend to be clustered, indicating that representative codes are more likely to be selected under our global updating strategy. In contrast, high-dimensional codes are highly informative due to their sparsity in the representation space, allowing for more even selection during training, which ensures high utilization with better performance. \textbf{3) Model Size}: \cref{tab:model_size} reveals that by increasing the number of ResBlock both in both the encoder and decoder, a higher reconstruction performance can be guaranteed. 

With these factors, we realize a super large codebook of 262,144 codebook size and 256 dimensions with high codebook usage (84\%), achieving the state-of-the-art reconstruction performance (1.00 rFID). To better illustrate the scaling properties, we also provide visualizations in \cref{fig:scaling_visualization}.

\begin{figure}[t]
    \centering
    \includegraphics[width=1.0\linewidth]{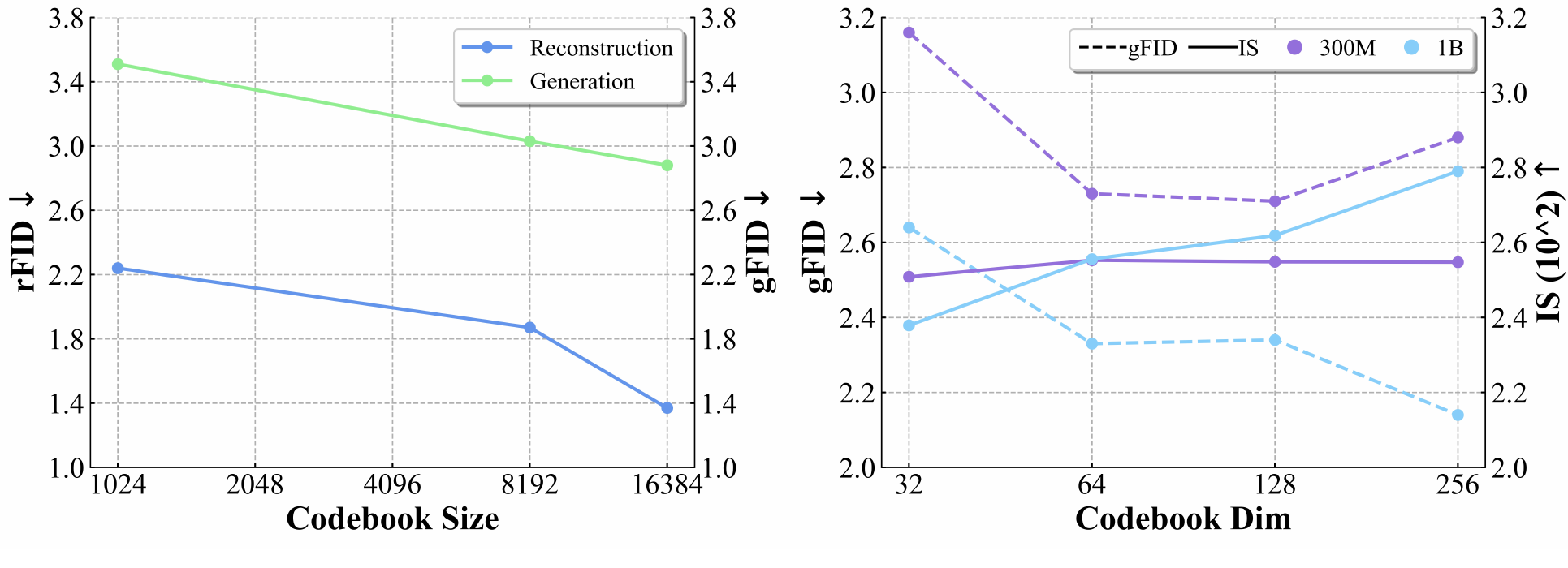}
    \vspace{-7mm}
    \caption{\textbf{Scaling up IBQ tokenizers enables better generation, especially with larger autoregressive models (\eg, 1B param.).} }
    \label{fig:scaling_recon_gen}
    \vspace{-6.5mm}
\end{figure}

\noindent \textbf{Scaling Up Tokenizers Improves Generation.} Scaling up IBQ tokenizers also enhances generation quality. As shown in \cref{fig:scaling_recon_gen}, increasing the codebook size significantly improves reconstruction and generation FID, with a similar trend observed when scaling code dimensions. Moreover, with larger autoregressive models (\eg, 1B parameters), the improvement in generation quality becomes more remarkable, suggesting that scaling up generative models can further unlock the potential of IBQ tokenizers.

\subsection{Ablation Studies}
\paragraph{Key Designs.} 
To validate the effectiveness of our method, we conduct ablation studies on several key designs, as shown in \cref{tab:module_ablation}. The re-implemented VQGAN performance is 3.98 rFID and 5.3\% codebook utilization. Different from previous methods, the replacement from VQ to IBQ achieves consistent distribution between encoded features and the whole codebook by rendering all code differentiable, which brings a clear improvement of both codebook usage (5.3\%$\rightarrow$ 98\%) and reconstruction quality (3.98 rFID$\rightarrow$1.67 rFID). By incorporating double quantization loss to force the selected code embeddings and encoded visual features toward each other, IBQ guarantees more precise quantization. Following MAGVIT-v2~\cite{magvit2}, we enlarge the model size for better compacity, and the reconstruction performance gets improved correspondingly.

\vspace{-5mm}
\paragraph{Comparison with LFQ.} 
For fair comparisons, we adopt LFQ~\cite{open-magvit2} with 16,384 codes and replace its asymmetric token factorization with our vanilla transformer architecture. \cref{tab:lfq_compare} shows that IBQ outperforms LFQ in both reconstruction and generation, which demonstrates increasing code dimension can improve the reconstruction ability of the visual tokenizer and further boost the visual generation. 

%% file: table/generation.tex
\begin{table*}[t]
    \centering
    \footnotesize
    \setlength{\tabcolsep}{12pt}
    \renewcommand\arraystretch{1.0}
    \resizebox{0.92\linewidth}{!}{
    \begin{tabular}{clccccc}
    \toprule
    \textbf{Type} & \textbf{Model} & \textbf{\#Para.} & \textbf{FID}$\downarrow$ & \textbf{IS}$\uparrow$ & \textbf{Precision}$\uparrow$ & \textbf{Recall}$\uparrow$  \\
    \midrule
    \multirow{4}{*}{\textcolor{gray}{Diffusion}} & \textcolor{gray}{ADM}~\citep{adm}  & \textcolor{gray}{554M}       & \textcolor{gray}{10.94} & \textcolor{gray}{101.0}        & \textcolor{gray}{0.69} & \textcolor{gray}{0.63}    \\
     & \textcolor{gray}{CDM}~\citep{cdm}   & \textcolor{gray}{$-$}       & \textcolor{gray}{4.88}  & \textcolor{gray}{158.7}       & \textcolor{gray}{$-$}  & \textcolor{gray}{$-$}   \\
     & \textcolor{gray}{LDM-4}~\citep{ldm} & \textcolor{gray}{400M}     & \textcolor{gray}{3.60}  & \textcolor{gray}{247.7}       & \textcolor{gray}{$-$}  & \textcolor{gray}{$-$}  \\
     & \textcolor{gray}{DiT-XL/2}~\citep{dit}  & \textcolor{gray}{675M}  & \textcolor{gray}{2.27}  & \textcolor{gray}{278.2}       & \textcolor{gray}{0.83} & \textcolor{gray}{0.57}   \\
    \midrule
    \multirow{4}{*}{VAR} & VAR-d16~\citep{var} & 310M & 3.30 & 274.4 & 0.84 & 0.51 \\
    & VAR-d20~\citep{var} & 600M & 2.57 & 302.6 & 0.83 & 0.56 \\
    & VAR-d24~\citep{var} & 1.0B & 2.09 & 312.9 & 0.82 & 0.59 \\
    & VAR-d30~\citep{var} & 2.0B & 1.92 & 323.1 & 0.82 & 0.59 \\
    \midrule
    \multirow{3}{*}{MAR} & MAR-B~\cite{mar} & 208M & 2.31 & 281.7 & 0.82 & 0.57 \\
    & MAR-L~\cite{mar} & 479M & 1.78 & 296.0 & 0.81 & 0.60 \\
    & MAR-H~\cite{mar} & 943M & \textbf{1.55} & 303.7 & 0.81 & \textbf{0.62}\\
    \midrule
    \multirow{21}{*}{Vanilla AR} & VQGAN~\citep{vqgan} & 227M & 18.65 & 80.4         & 0.78 & 0.26    \\
     & VQGAN~\citep{vqgan}    & 1.4B   & 15.78 & 74.3   & $-$  & $-$     \\
     & VQGAN-re~\citep{vqgan}  & 1.4B  & 5.20  & 280.3  & $-$  & $-$     \\
     & ViT-VQGAN~\citep{vit-vqgan} & 1.7B & 4.17  & 175.1  & $-$  & $-$        \\
     & ViT-VQGAN-re~\citep{vit-vqgan}& 1.7B  & 3.04  & 227.4  & $-$  & $-$     \\
     & RQTran.~\citep{rqvae}       & 3.8B  & 7.55  & 134.0  & $-$  & $-$     \\
     & RQTran.-re~\citep{rqvae}    & 3.8B & 3.80  & \textbf{323.7}  & $-$  & $-$    \\
     & LlamaGen-L~\citep{llamagen} & 343M & 3.80 & 248.28 & 0.83 & 0.51 \\
    & LlamaGen-XL~\citep{llamagen} & 775M & 3.39 & 227.08 & 0.81 & 0.54 \\
    & LlamaGen-XXL~\citep{llamagen} & 1.4B & 3.09 & 253.61 & 0.83 & 0.53 \\
    & LlamaGen-3B~\citep{llamagen} & 3.1B & 3.06 & 279.72 & 0.84 & 0.53 \\ 
    & LlamaGen-L$^{*}$~\citep{llamagen} & 343M & 3.07 & 256.06 & 0.83 & 0.52 \\
    & LlamaGen-XL$^{*}$~\citep{llamagen} & 775M & 2.62 & 244.08 & 0.80 & 0.57 \\
    & LlamaGen-XXL$^{*}$~\citep{llamagen} & 1.4B & 2.34 & 253.90 & 0.80 & 0.59 \\
    & LlamaGen-3B$^{*}$~\citep{llamagen} & 3.1B & 2.18 & 263.33 & 0.81 & 0.58 \\
    
     & Open-MAGVIT2-B~\cite{open-magvit2} & 343M & 3.08 & 258.26 & \textbf{0.85} & 0.51 \\
     & Open-MAGVIT2-L~\cite{open-magvit2} & 804M & 2.51 & 271.70 & 0.84 & 0.54 \\
     & Open-MAGVIT2-XL~\cite{open-magvit2} & 1.5B & 2.33 & 271.77 & 0.84 & 0.54 \\
     \midrule
     \multirow{4}{*}{Vanilla AR} & IBQ-B (Ours) & 342M & 2.88 & 254.73 & 0.84 & 0.51 \\
     & IBQ-L (Ours) & 649M & 2.45 & 267.48 & 0.83 & 0.52 \\
     & IBQ-XL (Ours) & 1.1B & 2.14 & 278.99 & 0.83 & 0.56 \\
     & IBQ-XXL (Ours) & 2.1B & 2.05 & 286.73 & 0.83 & 0.57 \\
    \bottomrule
    \end{tabular}}
    \vspace{-2mm}
    \caption{\textbf{Class-conditional generation on $\boldsymbol{256 \times 256}$ ImageNet.} $*$ specifies the generated images are $384 \times 384$ and are resized to 256×256 for evaluation. The evaluation protocol and implementation are the same as ADM~\cite{adm}.}
    \label{tab:gen1}
    \vspace{-5mm}
\end{table*}

%% file: sec/5_conclusion.tex
\section{Conclusion}
\label{sec:conclusion}
In this paper, we identify the bottleneck in scaling tokenizers (\eg, codebook size), stemming from the partial-update strategy in current VQ methods, which progressively enlarge the distribution gap between encoded features and non-activated codes, eventually leading to codebook collapse. To address this challenge, we propose a simple yet effective vector quantization method, termed as Index Backpropagation Quantization \textbf{(IBQ)}, for scalable tokenizer training, which updates all codes by applying the straight-through estimator on the categorical distribution over visual features and all codebook embeddings, thereby maintaining consistent distribution between the entire codebook and encoded features. Experiments on ImageNet demonstrate that IBQ enables a high-utilization, large-scale visual tokenizer with improved performance in both reconstruction (1.00 rFID) and generation (2.05 gFID).

%% file: sec/X_suppl.tex
\clearpage
\section*{Appendix}
\appendix

\input{table/model_configuration}
\input{table/ablation_soft_vq}
\input{table/computation_costs}

\section{Autoregressive Model Configurations}
We show the detailed autoregressive model configurations and training settings in \cref{tab:model_configuration}. We scale up the autoregressive models from 300M to 2.1B parameters, following the scaling rules proposed in VAR~\cite{var}.
\vspace{-1mm}

\section{Comparison with Soft Vector Quantization}

To comprehensively illustrate the rationality of our IBQ, we compare it with another global update method, Soft Vector Quantization (Soft VQ). During training, it adopts the weighted average of all code embeddings as the quantized feature $v_q$ and incorporates a cosine decay schedule of the temperature ranging from $0.9$ to $1e$-$6$ for one-hot vector approximation. As for inference, it switches back to the original VQGAN way, which selects the code with the highest probability for hard quantization. 

As shown in \cref{tab:ablation_soft_vq}, Soft VQ is far behind IBQ in both reconstruction quality and codebook usage. In the experiments, we observe that the training process of Soft VQ corrupts within a few epochs ($<10$). This may stem from the unstable adversarial training where the adaptive weight of the GAN loss appears enormous and ends up with NAN. In addition, the soft-to-hard manner for one-hot vector approximation brings more difficulty in optimization and incurs inconsistency of quantization between training and inference, as demonstrated by a significant reconstruction quality drop ($16.17\text{rFID}\rightarrow233.17\text{rFID}$). 

\begin{figure}[t]
    \centering
    \includegraphics[width=1.0\linewidth]{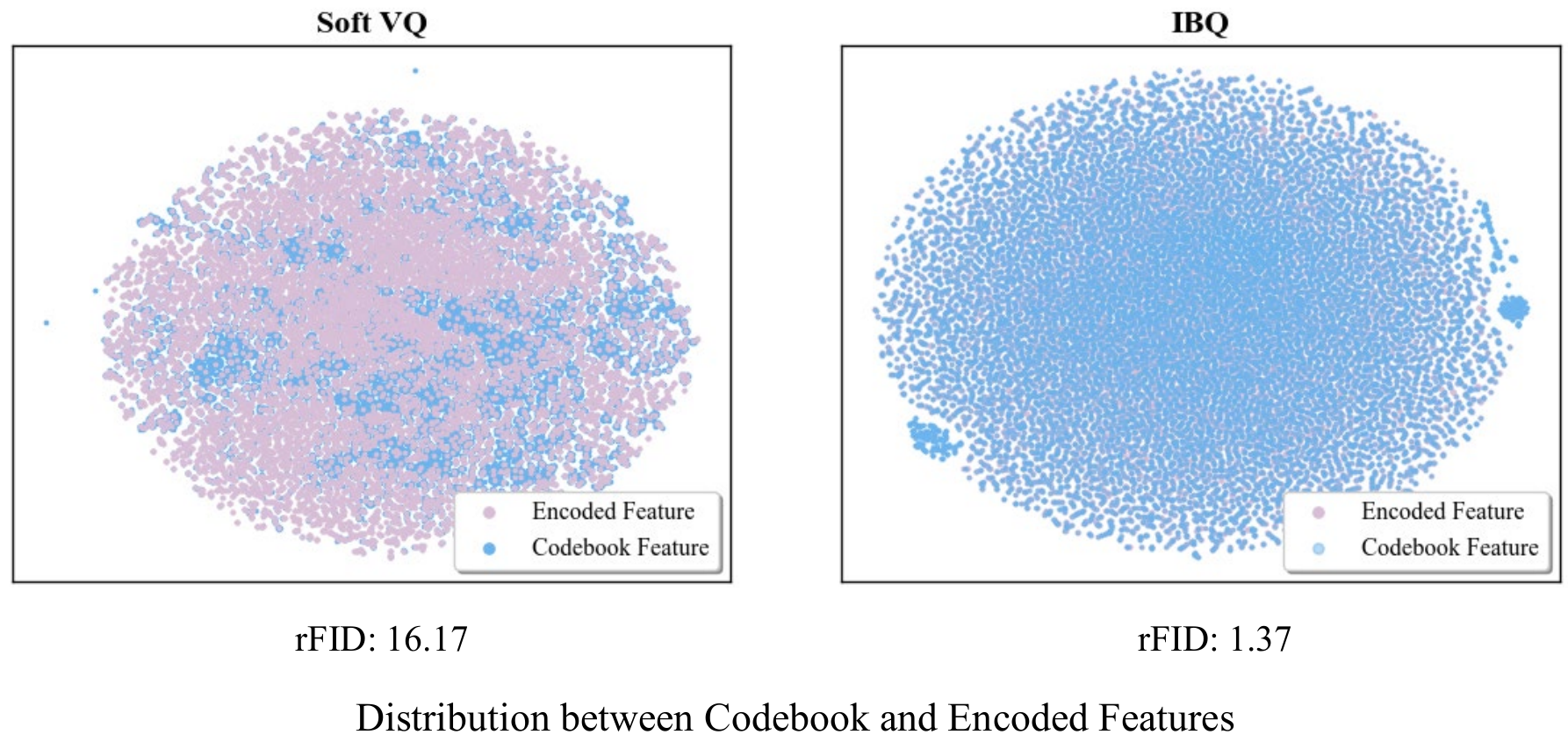}
    \vspace{-6mm}
    \caption{\textbf{Distribution Gap.} The T-SNE results of the codebook (16,384 codebook size and 256 dimension) and sampled encoded features.}
    \label{fig:distribution_gap_soft}
    \vspace{-5mm}
\end{figure}

Moreover, we provide an in-depth investigation by visualizing the distribution between the codebook and encoded features of Soft VQ. As shown in \cref{fig:distribution_gap_soft}, although all-code updating strategy is enabled, the inappropriate quantization process tends to cluster codes mistakenly, resulting in low codebook usage (2.5\%). We speculate that the force of the weighted average of code embeddings toward the encoded feature will smooth the codebook representation and result in similar and less informative code embeddings. In contrast, IBQ adopts hard quantization with index backpropagation. The hard quantization only involves the selected codes toward the encoded features for discriminative representation, thus ensuring precise quantization, while index backpropagation performs joint optimization of the entire codebook and visual encoder to achieve consistent distribution. Considering the factors above, our proposed IBQ shows dominance in both reconstruction quality and codebook utilization.

\begin{figure*}[t]
    \centering
    \includegraphics[width=\linewidth]{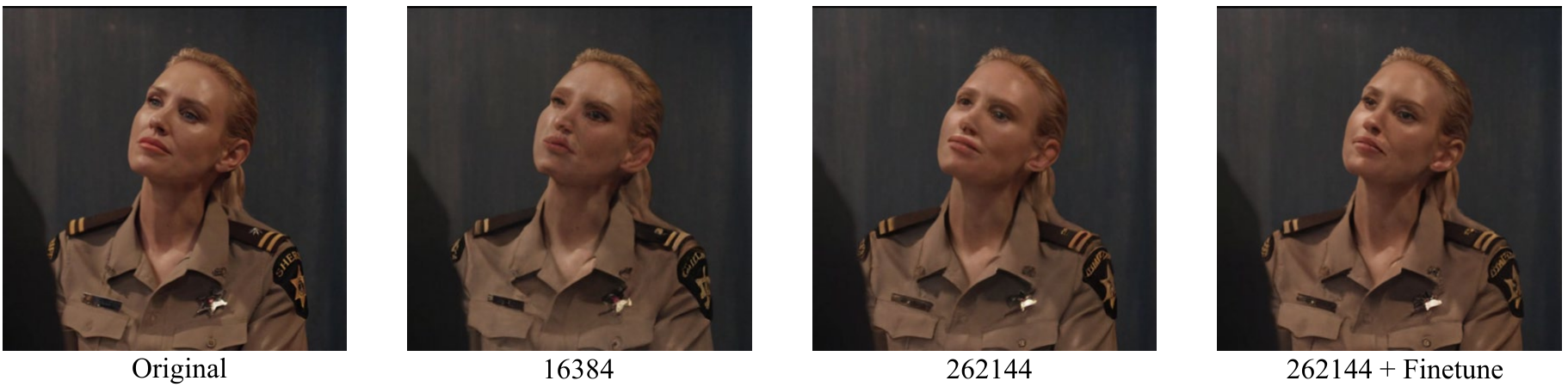}
    \caption{\textbf{Face reconstruction comparison.} Scaling up tokenizers and finetuning tokenizers on face data can effectively improve facial reconstruction performance.}
    \label{fig:face_recon}
\end{figure*}

\input{table/pretrain}

\section{Training Costs}
We evaluate the training costs of VQGAN and IBQ under varying codebook sizes using 8 A6000 GPUs. As shown in \cref{tab:computation_costs}, the all-codes updating mechanism of IBQ incurs only a marginal increase in training costs compared to VQGAN when the codebook size is up to 16,384, yet it significantly improves codebook utilization. Specifically, IBQ introduces an additional 0.2 GB of memory usage and extends training time by 19 minutes, but increases codebook utilization from 5.3\% to 96\%. Furthermore, VQGAN fails to train with an extremely large codebook (i.e., 262,144 entries), whereas IBQ successfully achieves 84\% utilization.

\section{Pretraining Tokenizer}
We further unveil the representation capacity of our tokenizer by pretraining IBQ on large-scale domain datasets, i.e., 1) General: CapFusion~\citep{capsfusion}, LAION-COCO~\citep{laion_coco}, CC12M~\citep{cc12m} and CC3M~\citep{cc3m}. 2) High-quality: LAION-aesthetics-12M\footnote{https://huggingface.co/datasets/dclure/laion-aesthetics-12m-umap}, LAION-aesthetics~\citep{laion_aesthetics}, JourneyDB~\citep{journeydb} and LAION-HD\footnote{https://huggingface.co/datasets/yuvalkirstain/laion-hd-subset}. We follow the same training settings stated in the manuscript while the training steps are $\sim$ 800,000. It can be seen in the \cref{tab:pretrain_recon} that IBQ achieves state-of-the-art performance compared to concurrent methods such as Cosmos~\cite{cosmos}, Show-o~\cite{showo}. Although some recent efforts in residual tokenization~\citep{tokenflow, infinity} can achieve better results, they are not listed here because residual techniques are orthogonal and compatible with IBQ. It is anticipated that our improvement on the naive quantization method better benefits the unified visual understanding and generation models compared to the residual one.

\section{Improving Face Reconstrution}
Visual tokenizers trained on ImageNet may not perform as expected for face reconstruction. Increasing the codebook size can effectively mitigate this limitation. As shown in \cref{fig:face_recon}, increasing the codebook size from 16,384 to 262,144 leads to improved face reconstruction quality. Additionally, incorporating face data into the training set or fine-tuning on face-specific datasets are effective strategies for further enhancement. In particular, fine-tuning IBQ on the FFHQ dataset further enhances reconstruction performance.

\section{Additional Visualizations}
\begin{figure*}[t]
    \centering
    \includegraphics[width=1.0\linewidth]{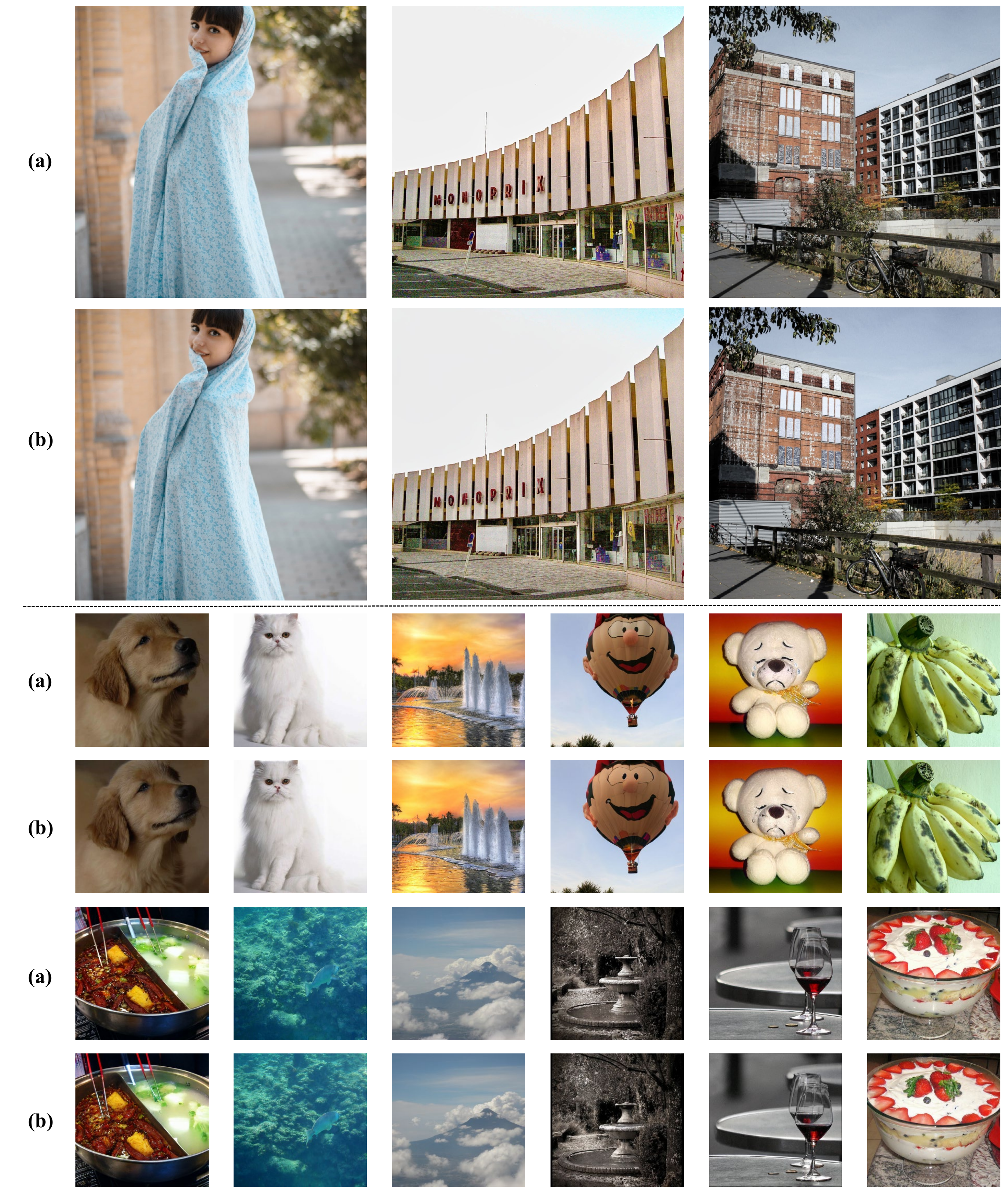}
    \caption{\textbf{Reconstruction samples.} The upper part illustrates the IBQ tokenizer tested at 1024 $\times$ 1024 Unsplash. While the second part showcases the IBQ tokenizer tested at 256 $\times$ 256 Imagenet. (a) indicates the original images and (b) signifies the reconstructions.}
    \label{fig:recon_case}
\end{figure*}

\begin{figure*}[t]
    \centering
    \includegraphics[width=1.0\linewidth]{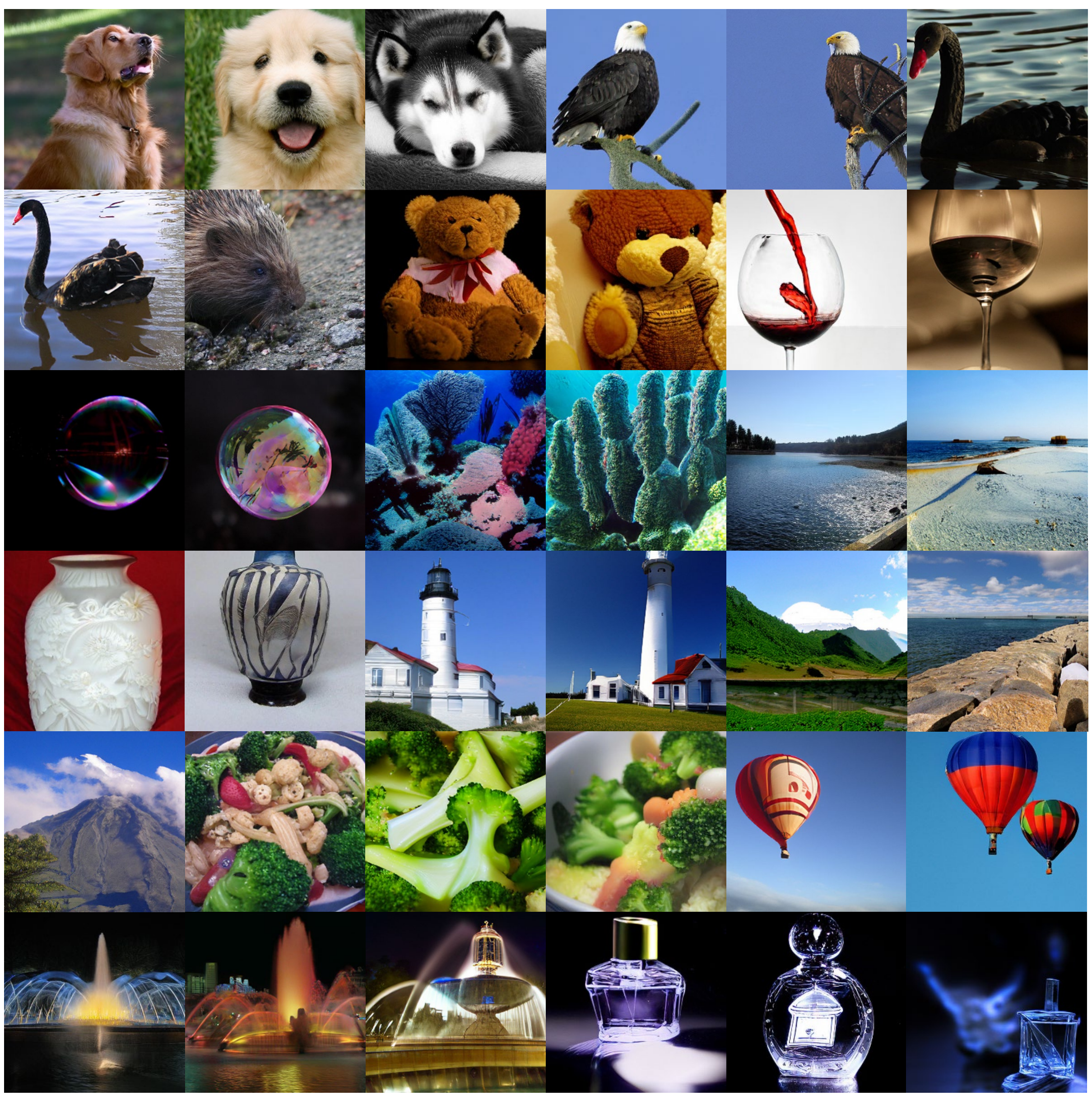}
    \caption{\textbf{Generation samples.} We showcase the 256 $\times$ 256 class conditional generation samples on Imagenet.}
    \label{fig:generation_case}
\end{figure*}

We provide more qualitative reconstruction and generation samples in \cref{fig:recon_case} and \cref{fig:generation_case}, respectively.
\vspace{-1mm}

%% file: table/model_configuration.tex
\begin{table}[t]
\centering
\resizebox{1.0\linewidth}{!}{
    \begin{tabular}{cccccccc}
    \toprule
    Model  & Parameters & Width $w$ & Head $h$ &  Depth $d$ & Lr & Batch Size & Epoch\\
    \midrule
    IBQ-B & 342M & 16 & 16 & 1024 & 3e-4 & 768 & 300 \\
    IBQ-L & 649M & 20 & 20 & 1280 & 3e-4 & 768 & 350 \\
    IBQ-XL & 1.1B & 24 & 24 & 1536 & 3e-4 & 768 & 400 \\
    IBQ-XXL & 2.1B & 30 & 30 & 1920 & 3e-4 & 768 & 450 \\
    \bottomrule
\end{tabular}
}
\vspace{-2mm}
\caption{\textbf{Model sizes and architecture configurations of IBQ.}}
\vspace{-2.5mm}
\label{tab:model_configuration}
\end{table}

%% file: table/ablation_soft_vq.tex
\begin{table}
\centering
\resizebox{\linewidth}{!}{
    \begin{tabular}{cccccc}
    \toprule
    Model & Optimization & Training & Inference & rFID$\downarrow$  & Usage$\uparrow$ \\
    \midrule
    Soft VQ & Corrupted & Soft & Soft & 16.17 & 2.5\%\\
    Soft VQ & Corrupted & Soft & Hard & 233.17 & 2.5\%\\
    IBQ (Ours)$^*$ & Stable & Hard & Hard  & 4.03 & 99\%\\
    IBQ (Ours) & Stable & Hard & Hard & 1.37 & 96\%\\
    \bottomrule
    \end{tabular}
}
    \vspace{-2mm}
    \caption{\textbf{Comparison with Soft Vector Quantization.} Soft VQ training corrupts after a few epochs. When adopting hard quantization for inference, there is a significant drop in rFID. $^*$ denotes IBQ with the same training epochs as Soft VQ.}
    \label{tab:ablation_soft_vq}
    \vspace{-2mm}
\end{table}

%% file: table/computation_costs.tex
\begin{table}[t]
\centering
\resizebox{1.0\linewidth}{!}{
\begin{tabular}{cccccc}
Model                  & Codebook Size & Parameters & Memory & Time/epoch & Usage \\
\toprule
\multirow{4}{*}{VQGAN} & 1,024	& 89.6M	& 19.5G	& 3h15min & 44\%\\
                       & 8,192	& 91.5M & 19.7G	& 3h18min & -\\
                       & 16,384 & 93.6M & 19.8G & 3h21min & 5.3\%\\
                       & 262,144 & 156M & 21.2G & 4h & $\sim$0\%\\
\midrule
\multirow{4}{*}{IBQ} & 1,024 & 89.6M & 19.5G & 3h20min &99\% \\
	            & 8,192	&91.5M	&19.7G	&3h30min	&98\% \\
	&16,384	&93.6M &20G	&3h40min &96\% \\
	&262,144 &156M &30.5G &9h &84\% \\

\bottomrule
\end{tabular}
}
\vspace{-2mm}
\caption{\textbf{Training computational costs comparison between VQGAN and IBQ.} (Tested on 8 A6000 gpus)}
\label{tab:computation_costs}
\vspace{-4mm}
\end{table}

%% file: table/pretrain.tex
\begin{table*}[t]
    \centering
    \setlength{\tabcolsep}{4pt}
    \renewcommand\arraystretch{1.1}
        \begin{tabular}{lcccccccccc}
        \toprule
        \multirow{2}{*}{\textbf{Method}} & \multirow{2}{*}{\textbf{Ratio}} & \textbf{Codebook} & 
        \multicolumn{3}{c}{\textbf{MS-COCO 2017}} & & \multicolumn{3}{c}{\textbf{Imagenet-1k}} \\
        \cmidrule{4-6} \cmidrule{8-10}
        & & \textbf{Size} & \textbf{rFID}$\downarrow$ & \textbf{PSNR}$\uparrow$ & \textbf{SSIM}$\uparrow$ & & \textbf{rFID}$\downarrow$ & \textbf{PSNR}$\uparrow$ & \textbf{SSIM}$\uparrow$ \\
        \midrule
        LlamaGen$^{\dagger}$ & 16 & 16384 & 8.40 & 20.28 & 0.55 & & 2.47 & 20.65 & 0.54 \\
        Show-o & 16 & 8192 & 9.26 & 20.90 & 0.59 & & 3.50 & 21.34 & 0.59 \\
        Cosmos & 16 & 64000 & 11.97 & 19.22 & 0.48 & & 4.57 & 19.93 & 0.49 \\ 
        Open-MAGVIT2 & 16 & 16384 & 7.93 & 22.21 & 0.62 & & 2.55 & 22.21 & 0.62 \\
        Open-MAGVIT2 & 16 & 262144 & 6.76 & 22.31 & 0.65 & & 1.67 & 22.70 & 0.64 \\
        IBQ (Ours) & 16 & 16384& 7.67 & 21.58 & 0.62 & & 2.06 & 22.01 & 0.61 \\
        IBQ (Ours) & 16 & 262144 & 6.79 & 22.28 & 0.65 & & 1.53 & 22.69 & 0.64 \\
    \bottomrule
    \end{tabular}
    \caption{Zero-shot reconstruction performance on ImageNet 50k validation set and MS-COCO val2017. The tokenizers are trained with large-scale general-domain datasets and aim to serve text-conditional image generation. The results are reported under the same setup for fair comparison (\textcolor{gray}{text in gray signifies the results directly from Cosmos report}). 
    $\dagger$ indicates that LlamaGen loads the model initially trained on Imagenet while the others are training from scratch, i.e., MS-COCO and Imagenet-1k are excluded from training data.}
    \label{tab:pretrain_recon}
\end{table*}

%% file: main.bbl
\begin{thebibliography}{52}
\providecommand{\natexlab}[1]{#1}
\providecommand{\url}[1]{\texttt{#1}}
\expandafter\ifx\csname urlstyle\endcsname\relax
  \providecommand{\doi}[1]{doi: #1}\else
  \providecommand{\doi}{doi: \begingroup \urlstyle{rm}\Url}\fi

\bibitem[Agarwal et~al.(2025)Agarwal, Ali, Bala, Balaji, Barker, Cai, Chattopadhyay, Chen, Cui, Ding, et~al.]{cosmos}
Niket Agarwal, Arslan Ali, Maciej Bala, Yogesh Balaji, Erik Barker, Tiffany Cai, Prithvijit Chattopadhyay, Yongxin Chen, Yin Cui, Yifan Ding, et~al.
\newblock Cosmos world foundation model platform for physical ai.
\newblock \emph{arXiv preprint arXiv:2501.03575}, 2025.

\bibitem[Baevski et~al.(2019)Baevski, Schneider, and Auli]{vq-wav2vec}
Alexei Baevski, Steffen Schneider, and Michael Auli.
\newblock vq-wav2vec: Self-supervised learning of discrete speech representations.
\newblock \emph{arXiv preprint arXiv:1910.05453}, 2019.

\bibitem[Bao et~al.(2021)Bao, Dong, Piao, and Wei]{beit}
Hangbo Bao, Li Dong, Songhao Piao, and Furu Wei.
\newblock Beit: Bert pre-training of image transformers.
\newblock \emph{arXiv preprint arXiv:2106.08254}, 2021.

\bibitem[Bengio et~al.(2013)Bengio, L{\'e}onard, and Courville]{ste}
Yoshua Bengio, Nicholas L{\'e}onard, and Aaron Courville.
\newblock Estimating or propagating gradients through stochastic neurons for conditional computation.
\newblock \emph{arXiv preprint arXiv:1308.3432}, 2013.

\bibitem[Borsos et~al.(2023)Borsos, Marinier, Vincent, Kharitonov, Pietquin, Sharifi, Roblek, Teboul, Grangier, Tagliasacchi, et~al.]{audiolm}
Zal{\'a}n Borsos, Rapha{\"e}l Marinier, Damien Vincent, Eugene Kharitonov, Olivier Pietquin, Matt Sharifi, Dominik Roblek, Olivier Teboul, David Grangier, Marco Tagliasacchi, et~al.
\newblock Audiolm: a language modeling approach to audio generation.
\newblock \emph{IEEE/ACM transactions on audio, speech, and language processing}, 31:\penalty0 2523--2533, 2023.

\bibitem[Brown et~al.(2020)Brown, Mann, Ryder, Subbiah, Kaplan, Dhariwal, Neelakantan, Shyam, Sastry, Askell, Agarwal, Herbert-Voss, Krueger, Henighan, Child, Ramesh, Ziegler, Wu, Winter, Hesse, Chen, Sigler, Litwin, Gray, Chess, Clark, Berner, McCandlish, Radford, Sutskever, and Amodei]{gpt3}
Tom Brown, Benjamin Mann, Nick Ryder, Melanie Subbiah, Jared~D Kaplan, Prafulla Dhariwal, Arvind Neelakantan, Pranav Shyam, Girish Sastry, Amanda Askell, Sandhini Agarwal, Ariel Herbert-Voss, Gretchen Krueger, Tom Henighan, Rewon Child, Aditya Ramesh, Daniel Ziegler, Jeffrey Wu, Clemens Winter, Chris Hesse, Mark Chen, Eric Sigler, Mateusz Litwin, Scott Gray, Benjamin Chess, Jack Clark, Christopher Berner, Sam McCandlish, Alec Radford, Ilya Sutskever, and Dario Amodei.
\newblock Language models are few-shot learners.
\newblock In \emph{NeurIPS}, pages 1877--1901, 2020.

\bibitem[Chang et~al.(2022)Chang, Zhang, Jiang, Liu, and Freeman]{maskgit}
Huiwen Chang, Han Zhang, Lu Jiang, Ce Liu, and William~T. Freeman.
\newblock Maskgit: Masked generative image transformer.
\newblock In \emph{CVPR}, pages 11305--11315, 2022.

\bibitem[Changpinyo et~al.(2021)Changpinyo, Sharma, Ding, and Soricut]{cc12m}
Soravit Changpinyo, Piyush Sharma, Nan Ding, and Radu Soricut.
\newblock Conceptual 12m: Pushing web-scale image-text pre-training to recognize long-tail visual concepts.
\newblock In \emph{CVPR}, pages 3558--3568, 2021.

\bibitem[Deng et~al.(2009)Deng, Dong, Socher, Li, Li, and Fei-Fei]{imagenet}
Jia Deng, Wei Dong, Richard Socher, Li-Jia Li, Kai Li, and Li Fei-Fei.
\newblock Image{N}et: A large-scale hierarchical image database.
\newblock In \emph{CVPR}, pages 248--255, 2009.

\bibitem[Dhariwal and Nichol(2021)]{adm}
Prafulla Dhariwal and Alexander Nichol.
\newblock Diffusion models beat gans on image synthesis.
\newblock \emph{NeurIPS}, 34:\penalty0 8780--8794, 2021.

\bibitem[Esser et~al.(2021)Esser, Rombach, and Ommer]{vqgan}
Patrick Esser, Robin Rombach, and Bj{\"{o}}rn Ommer.
\newblock Taming transformers for high-resolution image synthesis.
\newblock In \emph{CVPR}, pages 12873--12883, 2021.

\bibitem[Han et~al.(2024)Han, Liu, Jiang, Yan, Zhang, Yuan, Peng, and Liu]{infinity}
Jian Han, Jinlai Liu, Yi Jiang, Bin Yan, Yuqi Zhang, Zehuan Yuan, Bingyue Peng, and Xiaobing Liu.
\newblock Infinity: Scaling bitwise autoregressive modeling for high-resolution image synthesis.
\newblock \emph{arXiv preprint arXiv:2412.04431}, 2024.

\bibitem[Heusel et~al.(2017)Heusel, Ramsauer, Unterthiner, Nessler, and Hochreiter]{fid}
Martin Heusel, Hubert Ramsauer, Thomas Unterthiner, Bernhard Nessler, and Sepp Hochreiter.
\newblock {GAN}s trained by a two time-scale update rule converge to a local nash equilibrium.
\newblock In \emph{NeurIPS}, 2017.

\bibitem[Ho et~al.(2022)Ho, Saharia, Chan, Fleet, Norouzi, and Salimans]{cdm}
Jonathan Ho, Chitwan Saharia, William Chan, David~J Fleet, Mohammad Norouzi, and Tim Salimans.
\newblock Cascaded diffusion models for high fidelity image generation.
\newblock 23\penalty0 (1):\penalty0 2249--2281, 2022.

\bibitem[Huang et~al.(2023)Huang, Mao, Chen, and Zhang]{dqvae}
Mengqi Huang, Zhendong Mao, Zhuowei Chen, and Yongdong Zhang.
\newblock Towards accurate image coding: Improved autoregressive image generation with dynamic vector quantization.
\newblock In \emph{CVPR}, pages 22596--22605, 2023.

\bibitem[Isola et~al.(2017)Isola, Zhu, Zhou, and Efros]{patchgan}
Phillip Isola, Jun{-}Yan Zhu, Tinghui Zhou, and Alexei~A. Efros.
\newblock Image-to-image translation with conditional adversarial networks.
\newblock In \emph{CVPR}, pages 5967--5976, 2017.

\bibitem[Kingma(2013)]{vae}
Diederik~P Kingma.
\newblock Auto-encoding variational bayes.
\newblock \emph{arXiv preprint arXiv:1312.6114}, 2013.

\bibitem[Kingma(2014)]{adam}
Diederik~P Kingma.
\newblock Adam: A method for stochastic optimization.
\newblock \emph{arXiv preprint arXiv:1412.6980}, 2014.

\bibitem[Kynk{\"{a}}{\"{a}}nniemi et~al.(2019)Kynk{\"{a}}{\"{a}}nniemi, Karras, Laine, Lehtinen, and Aila]{precision_recall}
Tuomas Kynk{\"{a}}{\"{a}}nniemi, Tero Karras, Samuli Laine, Jaakko Lehtinen, and Timo Aila.
\newblock Improved precision and recall metric for assessing generative models.
\newblock In \emph{NeurIPS}, pages 3929--3938, 2019.

\bibitem[LAION(2022)]{laion_coco}
LAION.
\newblock Laion-coco 600m.
\newblock \url{https://laion.ai/blog/laion-coco}, 2022.

\bibitem[Lee et~al.(2022)Lee, Kim, Kim, Cho, and Han]{rqvae}
Doyup Lee, Chiheon Kim, Saehoon Kim, Minsu Cho, and Wook{-}Shin Han.
\newblock Autoregressive image generation using residual quantization.
\newblock In \emph{CVPR}, pages 11513--11522, 2022.

\bibitem[Li et~al.(2023)Li, Chang, Mishra, Zhang, Katabi, and Krishnan]{mage}
Tianhong Li, Huiwen Chang, Shlok Mishra, Han Zhang, Dina Katabi, and Dilip Krishnan.
\newblock Mage: Masked generative encoder to unify representation learning and image synthesis.
\newblock In \emph{CVPR}, pages 2142--2152, 2023.

\bibitem[Li et~al.(2024)Li, Tian, Li, Deng, and He]{mar}
Tianhong Li, Yonglong Tian, He Li, Mingyang Deng, and Kaiming He.
\newblock Autoregressive image generation without vector quantization.
\newblock \emph{arXiv preprint arXiv:2406.11838}, 2024.

\bibitem[Loshchilov(2017)]{adamw}
I Loshchilov.
\newblock Decoupled weight decay regularization.
\newblock \emph{arXiv preprint arXiv:1711.05101}, 2017.

\bibitem[Luo et~al.(2024)Luo, Shi, Ge, Yang, Wang, and Shan]{open-magvit2}
Zhuoyan Luo, Fengyuan Shi, Yixiao Ge, Yujiu Yang, Limin Wang, and Ying Shan.
\newblock Open-magvit2: An open-source project toward democratizing auto-regressive visual generation.
\newblock \emph{arXiv preprint arXiv:2409.04410}, 2024.

\bibitem[OpenAI(2023)]{openai2023gpt4}
OpenAI.
\newblock {GPT}-4 technical report.
\newblock \emph{arXiv preprint arXiv:2303.08774}, 2023.

\bibitem[Pan et~al.(2023)Pan, Sun, Ge, Li, Duan, Wu, Zhang, Zhou, Qin, Wang, et~al.]{journeydb}
Junting Pan, Keqiang Sun, Yuying Ge, Hao Li, Haodong Duan, Xiaoshi Wu, Renrui Zhang, Aojun Zhou, Zipeng Qin, Yi Wang, et~al.
\newblock Journeydb: A benchmark for generative image understanding.
\newblock \emph{arXiv preprint arXiv:2307.00716}, 2023.

\bibitem[Peebles and Xie(2023)]{dit}
William Peebles and Saining Xie.
\newblock Scalable diffusion models with transformers.
\newblock In \emph{CVPR}, pages 4195--4205, 2023.

\bibitem[Qu et~al.(2024)Qu, Zhang, Liu, Wang, Jiang, Gao, Ye, Du, Yuan, and Wu]{tokenflow}
Liao Qu, Huichao Zhang, Yiheng Liu, Xu Wang, Yi Jiang, Yiming Gao, Hu Ye, Daniel~K Du, Zehuan Yuan, and Xinglong Wu.
\newblock Tokenflow: Unified image tokenizer for multimodal understanding and generation.
\newblock \emph{arXiv preprint arXiv:2412.03069}, 2024.

\bibitem[Razavi et~al.(2019)Razavi, Van~den Oord, and Vinyals]{vqvae2}
Ali Razavi, Aaron Van~den Oord, and Oriol Vinyals.
\newblock Generating diverse high-fidelity images with vq-vae-2.
\newblock In \emph{NeurIPS}, 2019.

\bibitem[Rombach et~al.(2022)Rombach, Blattmann, Lorenz, Esser, and Ommer]{ldm}
Robin Rombach, Andreas Blattmann, Dominik Lorenz, Patrick Esser, and Bj{\"o}rn Ommer.
\newblock High-resolution image synthesis with latent diffusion models.
\newblock In \emph{CVPR}, pages 10684--10695, 2022.

\bibitem[Salimans et~al.(2016)Salimans, Goodfellow, Zaremba, Cheung, Radford, and Chen]{is}
Tim Salimans, Ian Goodfellow, Wojciech Zaremba, Vicki Cheung, Alec Radford, and Xi Chen.
\newblock Improved techniques for training gans.
\newblock In \emph{NeurIPS}, 2016.

\bibitem[Schuhmann and Beaumont(2022)]{laion_aesthetics}
Christoph Schuhmann and Romain Beaumont.
\newblock Laion-aesthetics.
\newblock \url{https://laion.ai/blog/laion-aesthetics/}, 2022.

\bibitem[Sharma et~al.(2018)Sharma, Ding, Goodman, and Soricut]{cc3m}
Piyush Sharma, Nan Ding, Sebastian Goodman, and Radu Soricut.
\newblock Conceptual captions: A cleaned, hypernymed, image alt-text dataset for automatic image captioning.
\newblock pages 2556--2565, 2018.

\bibitem[Shazeer(2020)]{glu}
Noam Shazeer.
\newblock Glu variants improve transformer.
\newblock \emph{arXiv preprint arXiv:2002.05202}, 2020.

\bibitem[Su et~al.(2024)Su, Ahmed, Lu, Pan, Bo, and Liu]{rope}
Jianlin Su, Murtadha Ahmed, Yu Lu, Shengfeng Pan, Wen Bo, and Yunfeng Liu.
\newblock Roformer: Enhanced transformer with rotary position embedding, 2024.

\bibitem[Sun et~al.(2024)Sun, Jiang, Chen, Zhang, Peng, Luo, and Yuan]{llamagen}
Peize Sun, Yi Jiang, Shoufa Chen, Shilong Zhang, Bingyue Peng, Ping Luo, and Zehuan Yuan.
\newblock Autoregressive model beats diffusion: Llama for scalable image generation.
\newblock \emph{arXiv preprint arXiv:2406.06525}, 2024.

\bibitem[Team(2024)]{chameleon}
Chameleon Team.
\newblock Chameleon: Mixed-modal early-fusion foundation models.
\newblock \emph{arXiv preprint arXiv:2405.09818}, 2024.

\bibitem[Tian et~al.(2024)Tian, Jiang, Yuan, Peng, and Wang]{var}
Keyu Tian, Yi Jiang, Zehuan Yuan, Bingyue Peng, and Liwei Wang.
\newblock Visual autoregressive modeling: Scalable image generation via next-scale prediction.
\newblock \emph{arXiv preprint arXiv:2404.02905}, 2024.

\bibitem[Touvron et~al.(2023)Touvron, Martin, Stone, Albert, Almahairi, Babaei, Bashlykov, Batra, Bhargava, Bhosale, Bikel, Blecher, Canton{-}Ferrer, Chen, Cucurull, Esiobu, Fernandes, Fu, Fu, Fuller, Gao, Goswami, Goyal, Hartshorn, Hosseini, Hou, Inan, Kardas, Kerkez, Khabsa, Kloumann, Korenev, Koura, Lachaux, Lavril, Lee, Liskovich, Lu, Mao, Martinet, Mihaylov, Mishra, Molybog, Nie, Poulton, Reizenstein, Rungta, Saladi, Schelten, Silva, Smith, Subramanian, Tan, Tang, Taylor, Williams, Kuan, Xu, Yan, Zarov, Zhang, Fan, Kambadur, Narang, Rodriguez, Stojnic, Edunov, and Scialom]{llama2}
Hugo Touvron, Louis Martin, Kevin Stone, Peter Albert, Amjad Almahairi, Yasmine Babaei, Nikolay Bashlykov, Soumya Batra, Prajjwal Bhargava, Shruti Bhosale, Dan Bikel, Lukas Blecher, Cristian Canton{-}Ferrer, Moya Chen, Guillem Cucurull, David Esiobu, Jude Fernandes, Jeremy Fu, Wenyin Fu, Brian Fuller, Cynthia Gao, Vedanuj Goswami, Naman Goyal, Anthony Hartshorn, Saghar Hosseini, Rui Hou, Hakan Inan, Marcin Kardas, Viktor Kerkez, Madian Khabsa, Isabel Kloumann, Artem Korenev, Punit~Singh Koura, Marie{-}Anne Lachaux, Thibaut Lavril, Jenya Lee, Diana Liskovich, Yinghai Lu, Yuning Mao, Xavier Martinet, Todor Mihaylov, Pushkar Mishra, Igor Molybog, Yixin Nie, Andrew Poulton, Jeremy Reizenstein, Rashi Rungta, Kalyan Saladi, Alan Schelten, Ruan Silva, Eric~Michael Smith, Ranjan Subramanian, Xiaoqing~Ellen Tan, Binh Tang, Ross Taylor, Adina Williams, Jian~Xiang Kuan, Puxin Xu, Zheng Yan, Iliyan Zarov, Yuchen Zhang, Angela Fan, Melanie Kambadur, Sharan Narang, Aur{\'{e}}lien Rodriguez, Robert Stojnic, Sergey Edunov,
  and Thomas Scialom.
\newblock Llama 2: Open foundation and fine-tuned chat models.
\newblock \emph{arXiv preprint arXiv:2307.09288}, 2023.

\bibitem[Tseng et~al.(2021)Tseng, Jiang, Liu, Yang, and Yang]{lecam}
Hung{-}Yu Tseng, Lu Jiang, Ce Liu, Ming{-}Hsuan Yang, and Weilong Yang.
\newblock Regularizing generative adversarial networks under limited data.
\newblock In \emph{CVPR}, pages 7921--7931, 2021.

\bibitem[Van Den~Oord et~al.(2017)Van Den~Oord, Vinyals, et~al.]{vqvae}
Aaron Van Den~Oord, Oriol Vinyals, et~al.
\newblock Neural discrete representation learning.
\newblock In \emph{NeurIPS}, 2017.

\bibitem[Vaswani et~al.(2017)Vaswani, Shazeer, Parmar, Uszkoreit, Jones, Gomez, Kaiser, and Polosukhin]{attention}
Ashish Vaswani, Noam Shazeer, Niki Parmar, Jakob Uszkoreit, Llion Jones, Aidan~N. Gomez, Lukasz Kaiser, and Illia Polosukhin.
\newblock Attention is all you need.
\newblock In \emph{NeurIPS}, pages 5998--6008, 2017.

\bibitem[Wang et~al.(2024)Wang, Zhang, Luo, Sun, Cui, Wang, Zhang, Wang, Li, Yu, et~al.]{emu3}
Xinlong Wang, Xiaosong Zhang, Zhengxiong Luo, Quan Sun, Yufeng Cui, Jinsheng Wang, Fan Zhang, Yueze Wang, Zhen Li, Qiying Yu, et~al.
\newblock Emu3: Next-token prediction is all you need.
\newblock \emph{arXiv preprint arXiv:2409.18869}, 2024.

\bibitem[Xie et~al.(2024)Xie, Mao, Bai, Zhang, Wang, Lin, Gu, Chen, Yang, and Shou]{showo}
Jinheng Xie, Weijia Mao, Zechen Bai, David~Junhao Zhang, Weihao Wang, Kevin~Qinghong Lin, Yuchao Gu, Zhijie Chen, Zhenheng Yang, and Mike~Zheng Shou.
\newblock Show-o: One single transformer to unify multimodal understanding and generation.
\newblock \emph{arXiv preprint arXiv:2408.12528}, 2024.

\bibitem[Yu et~al.(2022)Yu, Li, Koh, Zhang, Pang, Qin, Ku, Xu, Baldridge, and Wu]{vit-vqgan}
Jiahui Yu, Xin Li, Jing~Yu Koh, Han Zhang, Ruoming Pang, James Qin, Alexander Ku, Yuanzhong Xu, Jason Baldridge, and Yonghui Wu.
\newblock Vector-quantized image modeling with improved {VQGAN}.
\newblock In \emph{ICLR}, 2022.

\bibitem[Yu et~al.(2024{\natexlab{a}})Yu, Lezama, Gundavarapu, Versari, Sohn, Minnen, Cheng, Gupta, Gu, Hauptmann, Gong, Yang, Essa, Ross, and Jiang]{magvit2}
Lijun Yu, Jose Lezama, Nitesh~Bharadwaj Gundavarapu, Luca Versari, Kihyuk Sohn, David Minnen, Yong Cheng, Agrim Gupta, Xiuye Gu, Alexander~G Hauptmann, Boqing Gong, Ming-Hsuan Yang, Irfan Essa, David~A Ross, and Lu Jiang.
\newblock Language model beats diffusion - tokenizer is key to visual generation.
\newblock In \emph{ICLR}, 2024{\natexlab{a}}.

\bibitem[Yu et~al.(2024{\natexlab{b}})Yu, Sun, Zhang, Cui, Zhang, Cao, Wang, and Liu]{capsfusion}
Qiying Yu, Quan Sun, Xiaosong Zhang, Yufeng Cui, Fan Zhang, Yue Cao, Xinlong Wang, and Jingjing Liu.
\newblock Capsfusion: Rethinking image-text data at scale.
\newblock In \emph{Proceedings of the IEEE/CVF Conference on Computer Vision and Pattern Recognition}, pages 14022--14032, 2024{\natexlab{b}}.

\bibitem[Yu et~al.(2024{\natexlab{c}})Yu, Weber, Deng, Shen, Cremers, and Chen]{titok}
Qihang Yu, Mark Weber, Xueqing Deng, Xiaohui Shen, Daniel Cremers, and Liang-Chieh Chen.
\newblock An image is worth 32 tokens for reconstruction and generation.
\newblock \emph{arXiv preprint arXiv:2406.07550}, 2024{\natexlab{c}}.

\bibitem[Zhang et~al.(2018)Zhang, Isola, Efros, Shechtman, and Wang]{lpips}
Richard Zhang, Phillip Isola, Alexei~A. Efros, Eli Shechtman, and Oliver Wang.
\newblock The unreasonable effectiveness of deep features as a perceptual metric.
\newblock In \emph{CVPR}, pages 586--595, 2018.

\bibitem[Zhang et~al.(2022)Zhang, Roller, Goyal, Artetxe, Chen, Chen, Dewan, Diab, Li, Lin, Mihaylov, Ott, Shleifer, Shuster, Simig, Koura, Sridhar, Wang, and Zettlemoyer]{rmsnorm}
Susan Zhang, Stephen Roller, Naman Goyal, Mikel Artetxe, Moya Chen, Shuohui Chen, Christopher Dewan, Mona~T. Diab, Xian Li, Xi~Victoria Lin, Todor Mihaylov, Myle Ott, Sam Shleifer, Kurt Shuster, Daniel Simig, Punit~Singh Koura, Anjali Sridhar, Tianlu Wang, and Luke Zettlemoyer.
\newblock {OPT:} open pre-trained transformer language models.
\newblock \emph{arXiv preprint arXiv:2205.01068}, 2022.

\bibitem[Zhu et~al.(2024)Zhu, Wei, Lu, and Chen]{vqganlc}
Lei Zhu, Fangyun Wei, Yanye Lu, and Dong Chen.
\newblock Scaling the codebook size of vqgan to 100,000 with a utilization rate of 99\%.
\newblock \emph{arXiv preprint arXiv:2406.11837}, 2024.

\end{thebibliography}
